\documentclass[10pt,journal,compsoc]{IEEEtran}
\ifCLASSOPTIONcompsoc
  \usepackage[nocompress]{cite}
\else
  \usepackage{cite}

\fi
\ifCLASSINFOpdf
\else
\fi

\usepackage{times}
\usepackage{epsfig}
\usepackage{graphicx}
\usepackage{amsmath}
\usepackage{amssymb}
\usepackage{graphicx}
\graphicspath{ {images/} }

\usepackage{graphicx}
\usepackage{epsfig}
\usepackage{algorithm}
\usepackage{algorithmicx}
\usepackage{algpseudocode}
\usepackage{multirow}
\usepackage{lipsum}
\usepackage{graphicx}
\usepackage{amsmath,amssymb}
\usepackage{caption}
\usepackage{mathtools}
\usepackage{tikz}
\usepackage[breaklinks=true,bookmarks=false]{hyperref}

\usetikzlibrary{arrows,shapes,positioning,shadows,trees,calc,decorations.markings}

\tikzset{
  basic/.style   = {draw, text width=3cm, drop shadow, font=\sffamily, rectangle},
  root/.style    = {basic, rounded corners=2pt, thin, align=center, fill=gray!30},
  level 2/.style = {basic, rounded corners=6pt, thin,align=center, fill=gray!50, text width=10em},
  level 3/.style = {basic, thin, align=left, fill=gray!10, text width=8em, node distance=1.0cm and 3cm}
}

\begin{document}
%
\title{Learning from Few Samples: A Survey}

\author{Nihar~Bendre, Hugo~Terashima~Mar\'in,~\IEEEmembership{Senior~Member,~IEEE}
        and~Peyman~Najafirad,~\IEEEmembership{Senior~Member,~IEEE}
\IEEEcompsocitemizethanks{\IEEEcompsocthanksitem Nihar Bendre is with the Department of Electrical and Computer Engineering, University of Texas at San Antonio, San Antonio, TX, 78249. USA. \protect\\
E-mail: nihar.bendre@utsa.edu
\IEEEcompsocthanksitem Hugo~Terashima~Marín is a Full Professor with School of Engineering and Science, Tecnológico de Monterrey, Av. Eugenio Garza Sada 2501 Sur Col. Tecnológico. Monterrey N. L., CP. 64849. Mexico. \protect\\
E-mail: terashima@tec.mx
\IEEEcompsocthanksitem Dr. Paul Rad is an Associate Professor with the Department of Information Systems and Cyber Security, cofounder and director of Open Cloud Institute (OCI), University of Texas at San Antonio, San Antonio, TX, 78249. USA. \protect\\
E-mail: peyman.najafirad@utsa.edu}}
\IEEEtitleabstractindextext{%
\begin{abstract}
\label{abstract}

Deep neural networks have been able to outperform humans in some cases like image recognition and image classification. However, with the emergence of various novel categories, the ability to continuously widen the learning capability of such networks from limited samples, still remains a challenge. Techniques like Meta-Learning and/or few-shot learning showed promising results, where they can learn or generalize to a novel category/task based on prior knowledge. In this paper, we perform a study of the existing few-shot meta-learning techniques in the computer vision domain based on their method and evaluation metrics. We provide a taxonomy for the techniques and categorize them as data-augmentation, embedding, optimization and semantics based learning for few-shot, one-shot and zero-shot settings. We then describe the seminal work done in each category and discuss their approach towards solving the predicament of learning from few samples. Lastly we provide a comparison of these techniques on the commonly used benchmark datasets: Omniglot, and \textit{Mini}Imagenet, along with a discussion towards the future direction of improving the performance of these techniques towards the final goal of outperforming humans.

 
\end{abstract}

\begin {IEEEkeywords}
Meta-Learning, Zero-shot Learning, One-shot Learning, Few-Shot Learning, $n$-shot Learning Low-shot Learning, Representation Learning, Survey.
\end{IEEEkeywords}}


\maketitle

\IEEEdisplaynontitleabstractindextext

%
\IEEEpeerreviewmaketitle

\section{Introduction}
\label{intro}
\IEEEPARstart{A}{rtificial} intelligence (AI) based systems are becoming a huge part of the human life whether be it personal or professional. We are surrounded by AI-based machines and applications which intend to make our life easier. For example, the automatic mail filtering (spam detection), suggesting shopping websites, social networking in smartphones, etc. \cite{vaswani2017attention,bordes2014question,krizhevsky2012imagenet,sainath2013deep}. This impressive progress has been possible due to the break-through success achieved by machine or deep learning models \cite{lecun2015deep}. Machine or deep learning occupies a big part of the AI domain. Deep Learning models are built over multiple layers of perceptrons combined with the ability to apply gradient-based optimization techniques. Two of the most common  applications of deep learning models are: computer vision (CV), where the goal is to teach machines how to see and perceive things like humans do; Natural Language Processing (NLP) and Natural Language Understanding (NLU), where the goal is to analyze and comprehend large amounts of natural language data. These deep learning models have achieved tremendous success in image recognition \cite{he2016deep,simonyan2014very,szegedy2015going}, speech recognition \cite{mikolov2011strategies,panwar2017deep,Hansen2018,Hansen2019,hinton2012deep}, natural language processing and understanding \cite{collobert2011natural,bordes2015large,jean2014using,sahba2018automatic,ebadi2019implicit}, video analytics \cite{bendre2020human,sahba2018image,wang2018video,karpathy2014large,2020arXiv200611371D}, cyber security \cite{de2019implementation,parra2020detecting,de2020driverless,silva2019cooperative,chacon2019deep,silva2020temporal,silva2020opportunities}. The most common approach towards machine and/or deep learning is supervised learning, where large number of data samples, towards a particular application, are collected along with their respective labels and formed as a dataset. This dataset is categorized into three parts: training, validation and testing. During the training phase, the model is fed the data from the training and validation sets along with their respective labels and based on back propagation and optimization, the model generalizes to a hypothesis. During testing phase, the testing data is fed to the model and based on the derived hypothesis, the model predicts the output class of the testing data samples. 



The ability to handle large amounts of data, thanks to the power of computational and modern systems \cite{stewart2015jetstream,towns2014xsede}, has been exceptional. Along with the advancements of various algorithms and models, deep learning has been able to match up to humans and in some cases outperform humans. AlphaGo \cite{silver2017mastering} an AI-based agent, trained without any human guidance was able to defeat the world champion of Go, an ancient board game considered to be 10x complicated than chess \cite{lapan2018deep}; In another example of a complex and strategical multiplayer game called DOTA, the AI-agent was able to defeat human players of DOTA \cite{pachockiopenai}; For the task of image recognition and classification models like ResNet \cite{he2016deep} and Inception \cite{szegedy2017inception,szegedy2016rethinking,ioffe2015batch} were able to achieve better performance than humans on the popular ImageNet dataset which consists of over 14 million images with over 1000 classes \cite{deng2009imagenet}.

One of the ultimate goal of AI is to match or outperform humans in any given task. To achieve this goal, it is imperative to have minimal dependency on large balanced labeled datasets. Current models achieving successful results in tackling tasks with tremendous amounts of labelled data, however, approaches for other large variety tasks where the labelled data is scarce (few samples only) the performance of the respective models drops significantly. It is unrealistic to expect large balanced datasets for any particular task because due to nature of various categories it is nearly impossible to keep up with the producing labelled data. Furthermore, generation of labelled datasets require resources like time, human efforts and can be financially expensive. On the other hand, humans can quickly learn new class or classes, like given a photo of a strange animal, it can easily identify the animal from a photo which consists of a variety of animals. Another advantage of humans over machines is the ability to learn new concepts or classes on the fly, whereas machines have to go through an expensive offline process of training and retraining the entire model repeatedly to learn new classes, provided, the availability of labelled data. Researchers and developers are motivated to bridge this gap between humans and machines. As a potential solution to this problem, we have seen an ever increasing work in the area of meta-learning \cite{thrun2012learning,wang2019generalizing,finn2017meta,schmidhuber1987evolutionary,wang2016learning,wang2016learning1, vilalta2002perspective,wang2017learning,munkhdalai2017meta,rusu2018meta,santoro2017simple}, few-shot learning \cite{koch2015siamese,bertinetto2016learning,garcia2017few, bengio2013representation}, low-shot learning \cite{gidaris2018dynamic,qi2018low,wang2017multi,chu2018learning}, zero-shot learning \cite{frome2013devise,akata2015label,akata2015evaluation,xian2016latent,norouzi2013zero,tsai2017improving,xian2018feature}, where the goal is to make the model generalize better to novel tasks consisting of few labelled samples.

\subsection{What is Few-shot Learning and Meta-Learning}

  In few-shot, low-shot, or $n$-shot learning (where $n$ is generally between 1 to 5), the basic idea is to train the model with large amount of data samples on multiple categories, and during testing, the model is provided with novel categories (also referred to as novel set) where there are multiple data samples for each category and generally the number of categories is limited to five. In meta-learning, the goal is to generalize or learn the learning process, where the models are trained on a particular task, and a function of a different classifier is used on the novel set. The objective is to find the best hyperparameters and model weights where the model can easily adapt to the novel tasks without over-fitting to a novel task. In meta-learning, there are two categories of optimizational running simultaneously: one which learns to the new task; another which trains the learner. In recent times, few shot learning and meta-learning techniques have garnered interest, thus becoming a hot topic for research, with a flurry of recent papers  \footnote{four paper in CVPR 2018 to twenty papers in CVPR 2019 on the topic of few few meta-learning)} \cite{duan2016rl}.

Early work in the area of meta-learning was done by Yoshua and Samy Bengio \cite{bengio1990learning} and Fei-Fei Li in few-shot learning \cite{fe2003bayesian}. Metric learning is one of the older techniques used, where the objective is to learn from the embedding space. Images are transformed to their embeddings and images for a particular category were observed to be in close cluster whereas images from different categories were observed to be far away. Another popular approach where the data is augmented which results in generation of more samples from the limited few samples available. Currently semantics based approaches are extensively researched upon where the classification is based solely on the name of the category and its attribute. This semantics based approach is inspired towards solving zero-shot learning applications.

\begin{center}
    
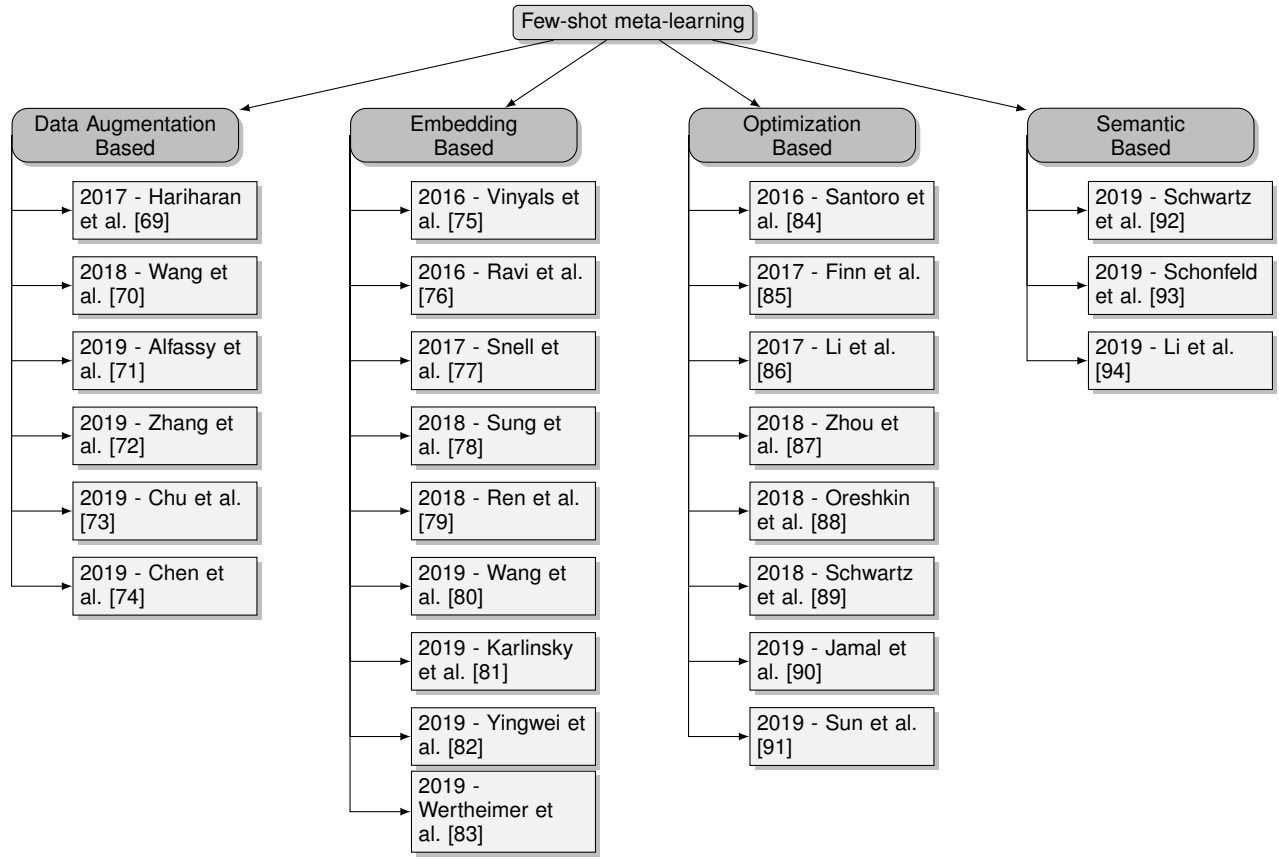
\begin{figure*}[t!]
    \centering
\begin{tikzpicture}[
  level 1/.style={sibling distance=45mm},
  edge from parent/.style={->,draw},
  >=latex]
  \footnotesize
\centering
\node[root] {Few-shot meta-learning}
  child {node[level 2] (c1) {Data Augmentation \\ Based}}
  child {node[level 2] (c2) {Embedding \\ Based}}
  child {node[level 2] (c3) {Optimization \\ Based}}
  child {node[level 2] (c4) {Semantic \\ Based}};

\begin{scope}[every node/.style={level 3}]

\node [below of = c1, xshift=15pt] (c12) {2017 - Hariharan et al. \cite{hariharan2017low}}; 
\node [below of = c12] (c13) {2018 - Wang et al. \cite{wang2018low}};
\node [below of = c13] (c14) {2019 - Alfassy et al. \cite{alfassy2019laso}}; 
\node [below of = c14] (c15) {2019 - Zhang et al. \cite{zhang2019few}};
\node [below of = c15] (c16) {2019 - Chu et al. \cite{chu2019spot}};
\node [below of = c16] (c17) {2019 - Chen et al. \cite{chen2019image}};


\node [below of = c2, xshift=15pt] (c21) {2016 - Vinyals et al. \cite{vinyals2016matching}};
\node [below of = c21] (c22) {2016 - Ravi et al. \cite{ravi2016optimization}};
\node [below of = c22] (c23) {2017 - Snell et al. \cite{snell2017prototypical}};
\node [below of = c23] (c24) {2018 - Sung et al. \cite{sung2018learning}};
\node [below of = c24] (c25) {2018 - Ren et al. \cite{ren2018meta}};
\node [below of = c25] (c26) {2019 - Wang et al.  \cite{wang2019tafe}};
\node [below of = c26] (c27) {2019 - Karlinsky et al. \cite{karlinsky2019repmet}};
\node [below of = c27] (c28) {2019 - Yingwei et al.  \cite{pan2019transferrable}};
\node [below of = c28] (c29) {2019 - Wertheimer et al. \cite{wertheimer2019few}};


\node [below of = c3, xshift=15pt] (c31) {2016 - Santoro et al. \cite{santoro2016one}};
\node [below of = c31] (c32) {2017 - Finn et al.  \cite{finn2017model}};
\node [below of = c32] (c33) {2017 - Li et al. \cite{li2017meta}};
\node [below of = c33] (c34) {2018 - Zhou et al.  \cite{zhou2018deep}};
\node [below of = c34] (c35) {2018 - Oreshkin et al. \cite{oreshkin2018tadam}};
\node [below of = c35] (c36) {2018 - Schwartz et al. \cite{schwartz2018delta}};
\node [below of = c36] (c37) {2019 - Jamal et al. \cite{jamal2019task}};
\node [below of = c37] (c38) {2019 - Sun et al.  \cite{sun2019meta}};

\node [below of = c4, xshift=15pt] (c41) {2019 - Schwartz et al. \cite{schwartz2019baby}};
\node [below of = c41] (c42) {2019 - Schonfeld et al. \cite{schonfeld2019generalized}};
\node [below of = c42] (c43) {2019 - Li et al. \cite{li2019large}};


\end{scope}
\foreach \value in {2,...,7}
  \draw[->] (c1.west) |- (c1\value.west);
\foreach \value in {1,...,9}
  \draw[->] (c2.west) |- (c2\value.west);
\foreach \value in {1,...,8}
  \draw[->] (c3.west) |- (c3\value.west);
\foreach \value in {1,...,3}
  \draw[->] (c4.west) |- (c4\value.west);

\end{tikzpicture}

\caption{\small{Taxonomy of Few-shot meta-learning techniques. We have classified these techniques under four categories based on data augmentation, embedding, optimization and semantic. Data augmentation based techniques involves approach where the limited data samples are augmented to generate more samples to enrich the training experience. Embedding based techniques involve approaches where the data is transformed to a low dimensional space and then clustered into different groups using a specific distance function. In optimization based techniques, a meta optimizer is used which learns/generalizes from the overall learning process. In semantic based learning, the semantic information of the data samples is used along with the samples to better generalize and thus predict the novel categories.}} 

\label{fig:tax_ml}
\end{figure*}
\end{center}

\subsection{Transfer Learning and Self-Supervised Learning}

The overall objective of transfer learning is to learn knowledge or experience from a set of tasks and transfer it to a task in the similar domain \cite{pan2009survey}. The task used to train the model to gain knowledge has lots of labelled samples whereas the transferred task has comparatively less labelled data (also called as fine-tuning), which is not enough to train and converge the model to the particular task. The performance of transfer learning technique is dependant on the relevance between the two tasks. While performing transfer learning, the classification layers are trained for the new tasks whereas the weights of the previous layers in a model are kept frozen \cite{yosinski2014transferable}. For every new task, where we do transfer learning, the choice of learning rate and the number of layers to be frozen has to be decided manually. In contrast to this, meta-learning techniques can quite rapidly adapt to a new task automatically. 

Self-supervised learning research have gained a lot of popularity in recent time \cite{larsson2016learning, zhang2016colorful, zhang2017split}. Training of Self-Supervised Learning (SSL) techniques is based on two steps: one, where the model is trained on a pre-defined pre-text task where it it trained on a large corpus of unlabelled data samples; two, where the learned model parameters are used to train or fine-tune the model for the main downstream task. The idea behind meta-learning or few-shot learning techniques is quite similar to self-supervised learning, which is to use prior knowledge, to recognize or fine tune to a novel task. Studies have shown that self supervised learning can be used along with few-shot learning to boost the performance of the model towards novel categories \cite{su2019does, jing2020self}. 

\begin{table}[b!]
\normalsize
\begin{tabular}{l|l}  \hline
\textbf{Symbol} & \textbf{Description}  \\ \hline
$D_{train}$ & Training set with limited number of samples \\ 
$D_{test}$ & Testing set   \\ 
$(x_n,y_n)$ & $n$ number of samples and their labels in $D_{train}$ \\
$(x^{\prime}_n,y^{\prime}_n)$ & $n$ number of samples and their labels in $D_{test}$ \\
$h$ & Original hypothesis \\ 
 $\hat{h}$ & Hypothesis for meta-learning \\ 
$\theta$ & model parameters \\ 
$(S)$ &  labelled support set \\
$(Q)$ & Query Set \\
$\mathcal{T}_i$ & $i$ Set of Tasks, where each task is a set of classes   \\ \hline

\end{tabular}
\caption{Table describing the various notations used throughout the paper.}
\label{table:notations}
\end{table}

\subsection{Taxonomy and Organization}

The main goal of meta-learning, few-shot learning, low-shot learning, one-shot, zero-shot learning, techniques is to make the deep learning model generalize to better to novel categories from handful samples with iterative training based on the prior knowledge or experience. Prior knowledge is knowledge acquired from training the samples on a labelled dataset consisting of large number of samples and then using this knowledge, THAT the model is trained ON, to recognize novel tasks with exposure to limited samples. Therefore, in the paper, we have combined all these techniques together under the main umbrella of  few-shot meta-learning. As there is no pre-define taxonomy to these techniques, we have classified these approaches into four main categories: data-augmentation based; metric-learning based, meta-optimization  based; and semantic-based (as illustrated in \autoref{fig:tax_ml}). Data-augmentation based techniques are quite popular where the idea is to expand the prior knowledge by augmenting the minimally available samples and generating more diverse samples to train the model. In embedding-based techniques,  the data samples are transformed to an another low-level dimension and then classified based on a distance between these embeddings. In optimization based techniques where a meta-optimizer is used to better generalize the model during the initial training and thus can do better prediction for the novel tasks. Semantic-based techniques are where the semantics of the data are used along with the prior knowledge for the model to either learn or optimize to novel categories. \autoref{table:notations} highlights the commonly used symbols used in various equations and algorithms in the rest of the paper along with their meaning. Our contributions to this paper are summarized as follows:

\begin{itemize}
    \item We analyze the seminal work done in the area of few shot meta-learning ( based on published research from the year 2016 to 2020) and provide a taxonomy and categorize the techniques into four categories: data augmentation based, embedding based, optimization based, and semantics based and summarize the work done in each of the proposed category.
    \item We did a comparison of the performance of the techniques in each category with reference to the two commonly used benchmark dataset: Onmiglot and \textit{Mini}Imagenet and discuss the limitations and possible future directions towards solving the problem of few shot meta-learning.
\end{itemize}

The remainder of the paper is structured as following. In \autoref{data_based} we discuss the data augmentation based techniques in which the training data is augmented to generate more samples to train the neural network model. In \autoref{data_based} the techniques where the input data samples are augmented to increase the training data are discussed. In \autoref{embedding_based}, we discuss the techniques where the general approach is to convert the high dimensional data to a lower dimension embedding and then using various distance or metric function compare the embedding of the initial trained tasks to that of the novel tasks. In \autoref{opti_based}, we discuss techniques in which a meta-optimizer is used which can learn from the training set and can generalize well on the novel tasks. In \autoref{semantic_based} we discuss techniques which extract semantic information from the training set to generalize on the novel tasks. In \autoref{discussion} we compare the performance of the various techniques discussed in the paper. This comparison is done on two benchmark datasets: Omniglot and \textit{mini}Imagenet.

\section{Few-Shot Meta Learning}
In this section, we describe the seminal work done in the data augmentation, embedding, optimization and semantic based learning approaches in the few shot meta learning domain as highlighted in \autoref{fig:tax_ml}. 
\subsection{Data Augmentation Based Techniques}
\label{data_based}

\begin{figure*}[t!]
\centering
\includegraphics[width=0.9\linewidth]{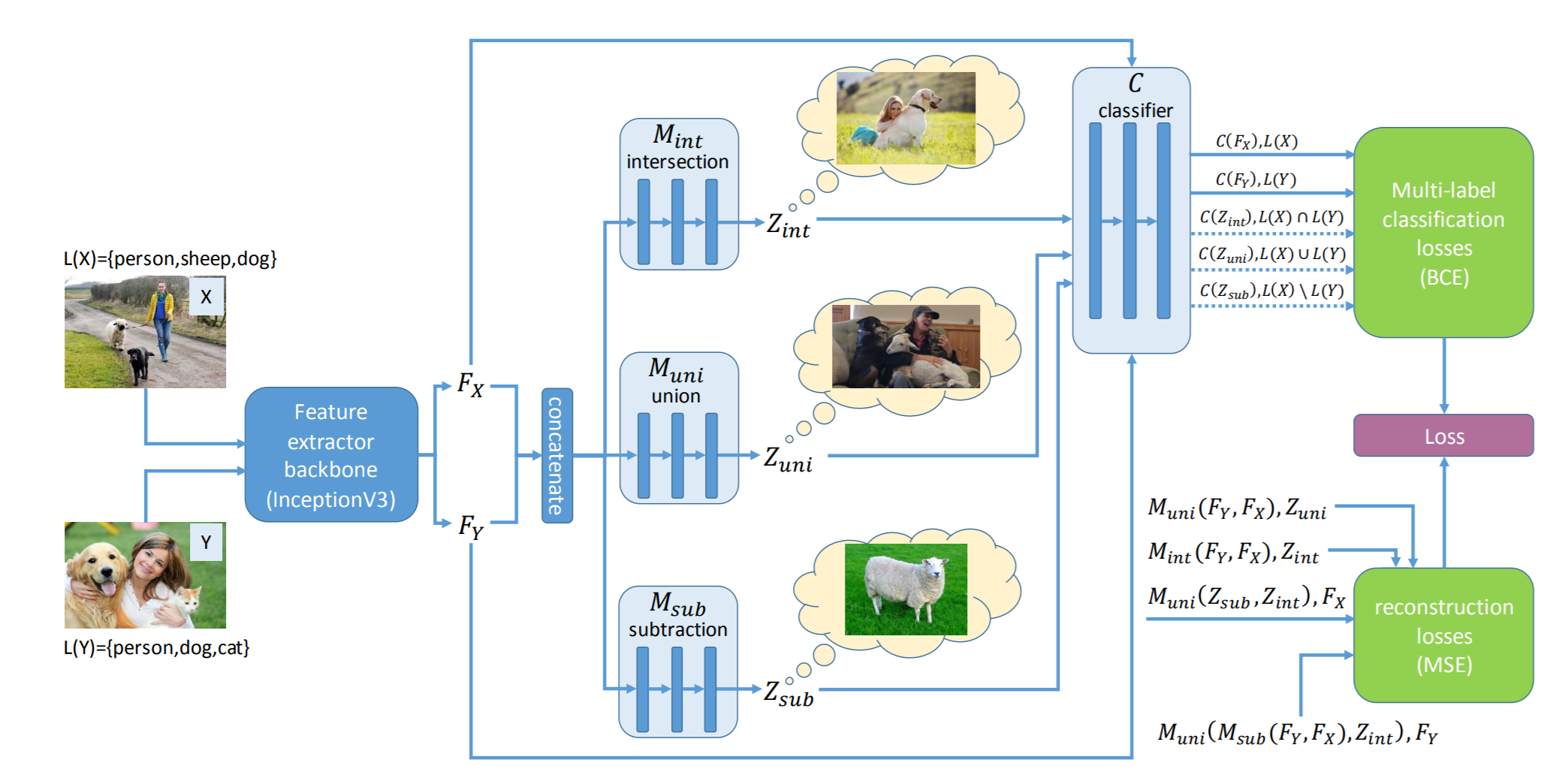}
\caption{The input the system are two different images, $x$ and $y$ with the set of their respective multiple labels $L(x), L(y)$. The labels in the feature space $F$ are represented with $F_x$ and $F_y$. The original vectors $L(x), L(y)$ along with the generated output of the three sub-modules are passed onto the classifier. Even though a set of pre-defined labels are used, the model can also generalize to labels which are not present into the set and are forced to learn implicitly only by observing the $L(x), L(y)$. Image Source: \cite{alfassy2019laso}}
\label{fig:laso}
\end{figure*}

Data-augmentation based techniques are quite popular in the supervised learning domain. Traditional techniques like scaling, cropping, rotating (clockwise and anti-clockwise) were implemented to expand the size of the training dataset where the goals was to make the model generalize better and avoid overfitting/underfitting scenarios. In the meta-learning space, the idea is to expand the prior knowledge by augmenting the minimally available samples and generating more diverse samples to train the model.

\subsubsection{LaSO: Label-Set Operations networks}

The work done by Alfassy et al. in \cite{alfassy2019laso} describes a technique to tackle samples with more than one label for few-shot classification settings. Their novel idea consists to combine multiple labels of samples in the feature space. The resultant feature vector will comprise of labels which have gone through a particular set of operations on the label set of the respective data sample-label pair. Using their method, data samples comprising of intersection, union, or set-difference which are generated from labels present in two different input data samples.

As illustrated in \autoref{fig:laso}, the input the system are two different images, $x$ and $y$ with the set of their respective multiple labels $L(x), L(y)$. The labels in the feature space $F$ are represented with $F_x$ and $F_y$. The Inception model \cite{szegedy2017inception} is used as backbone $\mathcal{B}$ to generate this feature space. The concatenated feature vectors, $F_x$ and $F_y$ are passed as an input to three sub-modules of the network $M_{int}, M_{uni}, M_{sub}$ which synthesize the feature vector in the same space. The original vectors $L(x), L(y)$ along with the generated output of the three sub-modules are passed onto the classifier. Even though a set of pre-defined labels are used, the model can also generalize to labels which are not present into the set and are forced to learn implicitly only by observing the $L(x), L(y)$. Code: \url{https://github.com/leokarlin/LaSO}\cite{alfassy2019laso}

\subsubsection{Recognition by Shrinking and Hallucinating Features}

In the work done by Hariharan and Girshick \cite{hariharan2017low}, the authors come up with a low-shot learning benchmark. This benchmark is inspired from the ImageNet1k dataset \cite{yalniz2019billion}. In the initial training step of the benchmark the learner is able to generalize to the $D_{train}$ from the dataset, and during the few shot training step, the model should be able to generalize from the feature space of the $D_{train}$ and the $D_{test}$, to correctly predict the novel tasks. The benchmark is used to provide a sanity check (by comparing the accuracy) on the model's ability to learn during its training on $D_{train}$ and testing with $D_{test}$. Code: \url{https://github.com/facebookresearch/low-shot-shrink-hallucinate}\cite{hariharan2017low}

They showed that by hallucinating the feature vector for the $D_{train}$, to train the model on more number of images. Doing so, it enhances the model's ability generalize better to novel class. The hallucination is done by using a $G$ function. The $G$ function consists of three fully connected MLP layers. For the model to learn and generalize better to the novel class, the authors have introduced a new loss function called the squared gradient magnitude loss (SGM) which is applied during the few-shot learning phase. The loss is given by:

\begin{equation}
\label{equ:hallu_1}
min \mathcal{L}_{D}(\theta, C) = min_{(c,\theta)} \frac{1}{|D|} \sum_{(x_n,y_n) \in D} L_{cls}(C, \theta(x_n),y_n)
\end{equation}

where, $C$ is the classifier which is used in the model. $D$ is $D_{train}$. $L_{cls}$ is given by:

\begin{equation}
\label{equ:hallu_2}
L_{cls} (C, x_n, y_n) = -log p_{y_n}(c,x_n)
\end{equation}

\subsubsection{Learning via Saliency-guided Hallucination}

The work done by Zhang et al. in \cite{zhang2019few} is based on using data hallucination technique in which they use a saliency network \cite{liu2016dhsnet, alter1998extracting} to generate the background and foreground information of an image. Using a two-stream network which generates hallucinated data points in the feature space based on the foreground and background information. Their model takes advantage of the generated saliency maps to improve the performance of few-shot technique.

Their model consists of three modules: one, Saliency Network; two, A network to encode and mix the foreground and background information (FEMN);  and three, A Similarity network. The saliency network generates the saliency maps based on the feature vector of the support $S$ samples. The FEMN combines the foreground and background information. The similarity network determines if the query image and and the samples from the support set. The data hallucination process is the summation of foreground and the background information. Code: \url{https://github.com/HongguangZhang/SalNet-cvpr19-master}\cite{zhang2019few}

\subsubsection{Low-Shot Learning from Imaginary Data}

The work done by Wang et al. \cite{wang2018low} is based on hallucinating the $D_{train}$ data samples which can be useful for the classifier to learn or generalize to novel tasks. Instead of using hallucination to generate more diverse data like in \cite{zhang2019few, hariharan2017low}, their goal is to generate hallucinated data which is related to the samples in $D_{train}$. In their approach, they introduce a hallucinator along with the meta-learner to learn and generalize to the novel tasks. The objective of the hallucinator model is to map the hallucinated data to the original samples in $D_{train}$. The hallucinator model produces an extended set of $D_{train}$ after being trained on the base $D_{train}$. Code: \url{https://github.com/facebookresearch/low-shot-shrink-hallucinate}\cite{wang2018low}

\subsubsection{A Maximum-Entropy Patch Sampler}

The work done by Chu et al. in \cite{chu2019spot} is based on a ``learned" form of data augmentation where they search from various sequences of patches generated by sampling the image and perform classification on those patches using the extracted features. They claim that their method, along with the positive and negative sampling rules, together with an improved reward function (based on Maximum Entropy Reinforcement Learning \cite{chu2019spot}), improves the performance for $n$-shot learning paradigm. 

Their model consists of five modules: feature extractor, a CNN based model, which at each time step extracts the feature embedding on the patches (generated by the maximum entropy sampler).; state encoder, which is used to aggregate the features generated by the feature extractor step. To achieve this, they use a RNN based GRU \cite{cho2014properties} model; a maximum entropy Sampler which generates patches from the input data sample. This is build by taking inspiration from Soft Q- Learning \cite{haarnoja2017reinforcement}; Action context encoder's goal is to take into consideration all the global information generated by the maximum entropy sampler along with the feature extractor modules; classifier, whose objective is to accurately differentiate the input data sample to generate it predicted label.

\subsubsection{Image Deformation Meta-Networks}

The work done by Chen et al. in \cite{chen2019image} generates additional training samples by combining a meta-learner module with an image deformation module. These aforementioned modules are trained in an end-to-end manner to significantly outperforms the then state-of-the-art technique. The meta-learning module learns from a group of $k-way, n-shot$ tasks to classify samples from $D_{train}$ which is then evaluated on samples from $D_{test}$. The deformation module adaptively fuses the images from the support set $S$ to generate synthesized deformed images which are then mapped to the feature vector generated from $D_{train}$ through an embedding sub-module to construct the one-shot classification. Code: \url{https://github.com/tankche1/IDeMe-Net}

\subsection{Embedding-Based Techniques}
\label{embedding_based}
\begin{figure}[t!]
\centering
\includegraphics[width=0.9\linewidth]{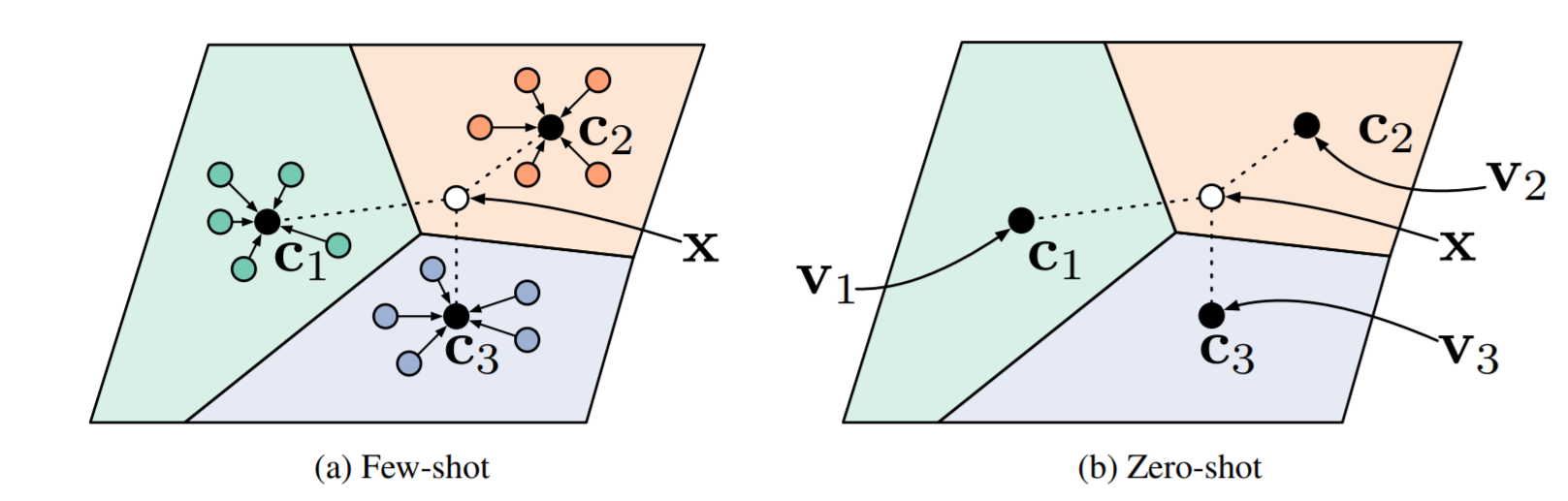}
\caption{Figure explaining the prototypical network for few-shot and zero-shot application, where $c_k$ are prototypes for few-shot learning application and $v_k$ is the meta-data. An embedding of query points for an image $\hat{x}$ is classified using a softmax function performed over the class function using squared Euclidean distance function. Image Source: \cite{snell2017prototypical}}
\label{fig:prototypical_network}
\end{figure}

In this subsection, we discuss techniques are which are model based, where the goal is to find best possible hypothesis from the hypothesis space and which can generalize well to a variety of tasks. Embedding based techniques also known as metric based techniques, is where the data is transformed to a lower dimension representation and then clustered and compared using a specific distance/metric function. 



\subsubsection{Relation Network}

The work done by Sung et al. \cite{sung2018learning} proposes a flexible and simple yet effective network called as Relation Network (RN), for the $n$-shot learning settings. The basic idea with RN is based on episode training \cite{vinyals2016matching}, where an episode consists of randomly selected tasks from the $D_{train}$ with $k$ number of labelled samples from each class. 

The relation network is based on two steps: one, where the samples from $D_{train}$ and the query set are transformed to a low level embedding space. This process is done by the embedding module; two, using a relation module these low level representations are compared and determined if the query image is matching to any of the output categories. The embedding module consists of four layers of convolutional neural network where each layer consists of 64 convolutional filters with a $3 \times 3$ kernel size, followed by batch normalization, ReLU activation and max-pooling performed using a $2 \times 2$ window. The relation module is a basic comparison module where the use mean square error (MSE) as a distance function. Code: \url{https://github.com/floodsung/LearningToCompare_FSL}

\subsubsection{Prototypical Network}


The work done Snell et al. in \cite{snell2017prototypical} describes a novel then state-of-the-art network called \textit{Prototypical Network} to target few-shot and zero-shot applications. The network computes a prototype representation (an $M$-dimensional representation) of each class using an embedding function $f_{\theta} : \mathbb{R}^D \rightarrow \mathbb{R}^M$, where $\theta$ is the learnable parameters. The prototype is given by $c_k \in \mathbb{R}^M$ which is calulated using the mean vector of the embedded support points in the class space. $c_k$ is given by: 

\begin{equation}
\label{equ:c_k}
c_k = \frac{1}{s_k} \sum\limits_{(x_n, y_n) \in s_k} f_{\theta} (x_n)
\end{equation}

where $s_k$ is the number of samples in true class $k$

The authors have primarily used squared Euclidean distance ($D$) for their prototypical network where where $D : \mathbb{R}^M \times \mathbb{R}^M \rightarrow [0, +\infty]$. For a querry $x^\prime$ the prototypical network, based on the softmax \cite{liu2016large} performed on the distances to the prototypes in embedding space produces a distribution over the total range of classes using the embedding class meta-data $v_k$. Code: \url{https://github.com/jakesnell/prototypical-networks}


\subsubsection{Learning in localization of realistic settings}

Wertheimer et al. \cite{wertheimer2019few} proposes an incremental work on top of prototypical network \cite{snell2017prototypical} to target the realistic open world images involve thousands of different classes with subtle variations. They target the problems of heavy class imbalance, heavy tailored and fine-grained clutter recognition. Their work is incremental on the existing work of prototypical networks that results in significant increase in the performance without adding much complexity to the over network. They propose a new training approach to tackle the class imbalance problem which is based on top of leave-one-out cross validation. To tackle the clutter problem, they use an learner architecture which can efficiently localize and object before classifying them into various classes. To tackle the fine-grain problem and to differentiate subtleties in an object, they use bilinear pooling \cite{carreira2012semantic, lin2015bilinear} to increase the representation power of the learner model. Using the combination of the three improvement techniques, Wertheimer et al. were able to double the results with respect to accuracy of prototypical networks on meta-iNat benchmark.

\textit{Cross Validation using Leave-one-Out Approach:} To have a successful prototypical network which can recognise novel rare classes, it needs to be trained on a relatively great size of images belonging to the common class and on the few referenced images belonging to the rare novel class. To achieve this, either the batch size needs to be increased or reduce the number of referenced categories during the training. Increasing the batch size is not always an option due to the computational limitations, whereas reducing the novel referenced images can result in the model learning poorly and the center of each class becoming distorted. To overcome this, their approach uses cross validation based on leave-one-out approach.

Let $c$ be the entire set of classes and $s_k$ is the number of samples in each class. $v_{n,k}$ is the feature vector of the $n^{th}$ sample in the $c^{th}$ category, the prototypes are generated in the following manner:

\begin{equation}
\label{equ:c_k1}
c_k = \frac{\sum\limits_{n=1}^N v_{n,k}}{s_k}  
\end{equation}

\textit{Effective Localization:} To distinguish relevant object from a clustered image is highly difficult when it is trained on a few images and their respective labels. To address this issue, the authors proposed to localize the object in the referenced image and the query image which can make the process of classification significantly better. They used two approaches to isolate the images: Unsupervised localization - where a category-agnostic learner model is internally developed on the $D_{train}$; and few-shot localization - where the images from $D_{train}$ are used to generate bounding box on the $D_{test}$.

The procedure for both the localization techniques is same where a sub-module of localizer is used to classify the location of every object in the final $10 \times 10$ feature map layer of the model and categorized as `foreground' and `background' predictions. A softmax function is applied on these prediction embeddings and using a L2 distance. The training is done end-to-end for both of the localization procedures. The localizer is trained and used only for the purpose of classification.

\textit{Bilinear pooling:} Techniques like fisher vectors \cite{perronnin2010improving} and others \cite{acharya2018covariance, jegou2010aggregating, csurka2004visual, gao2016compact} are used to increase the feature space $F$, and increase the fine-grain classification power of the models. These techniques are applied to the fully-connected layers or the classifier layers in a model which increases the model parameters and it complexity. The approach used by Wertheimer et al. of bilinear pooling \cite{lin2015bilinear} takes into account two feature maps and computes the cross variance amongst them and performs a pixel-wise product and then does average pooling on top of it. By doing so, they claim that no extra parameters are added to the network. Code: \url{https://github.com/daviswer/fewshotlocal}

\begin{figure}[t!]
\centering
\includegraphics[width=0.95\linewidth]{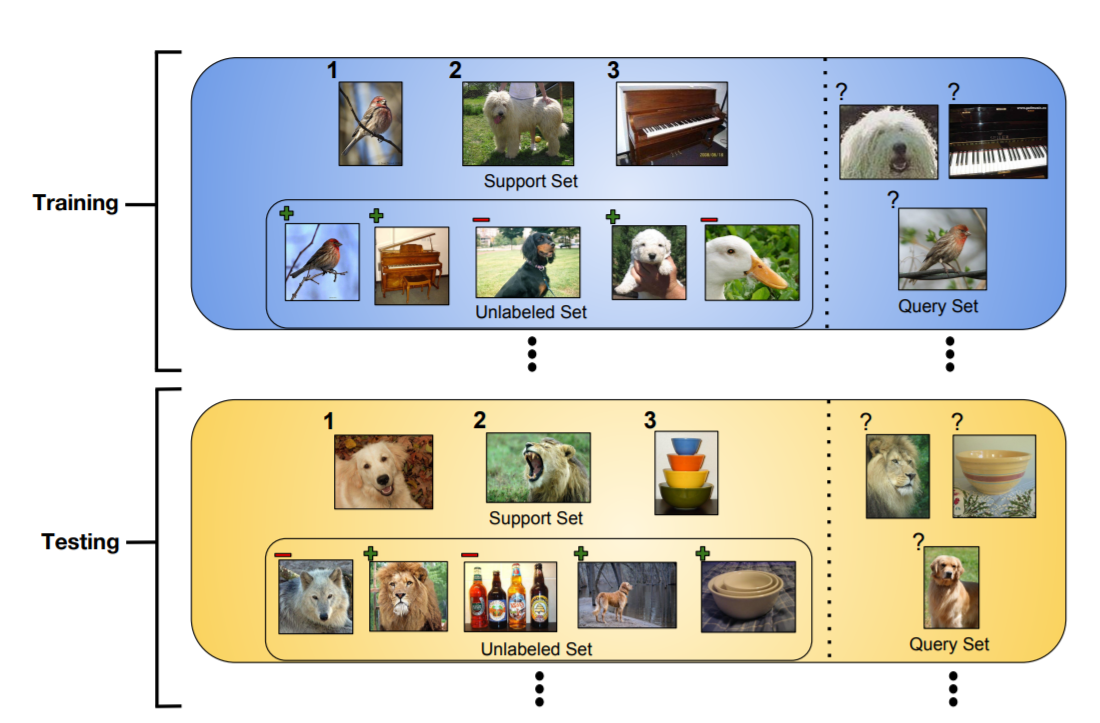}
\caption{Figure explaining the semi-supervised few-shot learning setup. $(S)$ and $(R)$ are sets of labelled and unlabelled data samples respectively and the goal is to use these samples to generalize well on the query set $(Q)$. The green plus sign indicates that the images in the distractor set that belong to the classes that of in the training set, whereas the ones indicated by red minus sign are the images in the distractor class that do not belong to classes in the training dataset. Source: \cite{ren2018meta}}
\label{fig:semi-supervised}
\end{figure}

\subsubsection{Learning for Semi-Supervised Classification}

In this work, Ren et al. \cite{ren2018meta} propose a semi-supervised and novel extension to the prototypical networks to deal with scenarios where unlabelled data samples are available along with $D_{train}$ samples and their respective labels can generate prototypes. As shown in \autoref{fig:semi-supervised}, they take into consideration two scenarios: one where the unlabelled data samples belong to the same set of classes $c$; and other where the unlabelled data samples belong to another set of classes called as distractor classes. In the semi-supervised based few-shot learning approach the $D_{train}$ consists of a tuple $(S, R)$ where $(S)$ is the set of labelled samples and $(R)$ is the set of unlabelled samples. $(S)$ is the reference set of support set used in the prototypical networks. This set consists of various images and their respective labels. 

\begin{figure}[b!]
\centering
\includegraphics[width=0.9\linewidth]{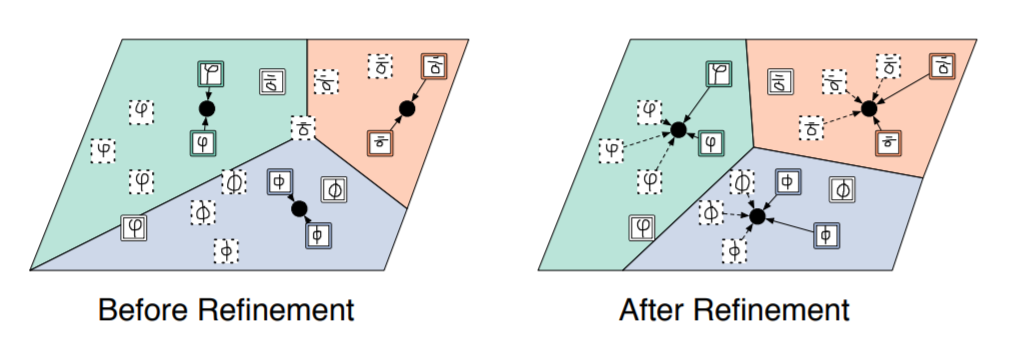}
\caption{Figure indicating the prototype clustering before and after refinement. Image Source: \cite{ren2018meta}}
\label{fig:b&f_refinement}
\end{figure}

If the original prototypical networks discussed in \cite{snell2017prototypical} is considered, it can successfully generate prototypes $c_k$ for the labelled set $(S)$ but fails to generate prototypes $(\hat{c_k})$ for the unlabelled set $(R)$. The authors provide various techniques to process on the labelled set $(S)$ and generate refined prototypes for the unlabelled $(R)$ set. One the refined prototypes $(\hat{c_k})$ for $(R)$ are generated, the model is trained with the same loss function as used for the vanilla prototypical network (refer \autoref{equ:c_k}) for the refined prototypes $(\hat{c_k})$. After that each query is classified based on the distance function and the proximity of the query to the generated refined prototypes $(\hat{c_k})$ using average negative log probability (refer \autoref{fig:b&f_refinement}. The refined prototypes, which are generated by considering the samples from $(R)$, are seen to be classified accurately.

\begin{figure*}[t!]
\centering
\includegraphics[width=0.9\linewidth]{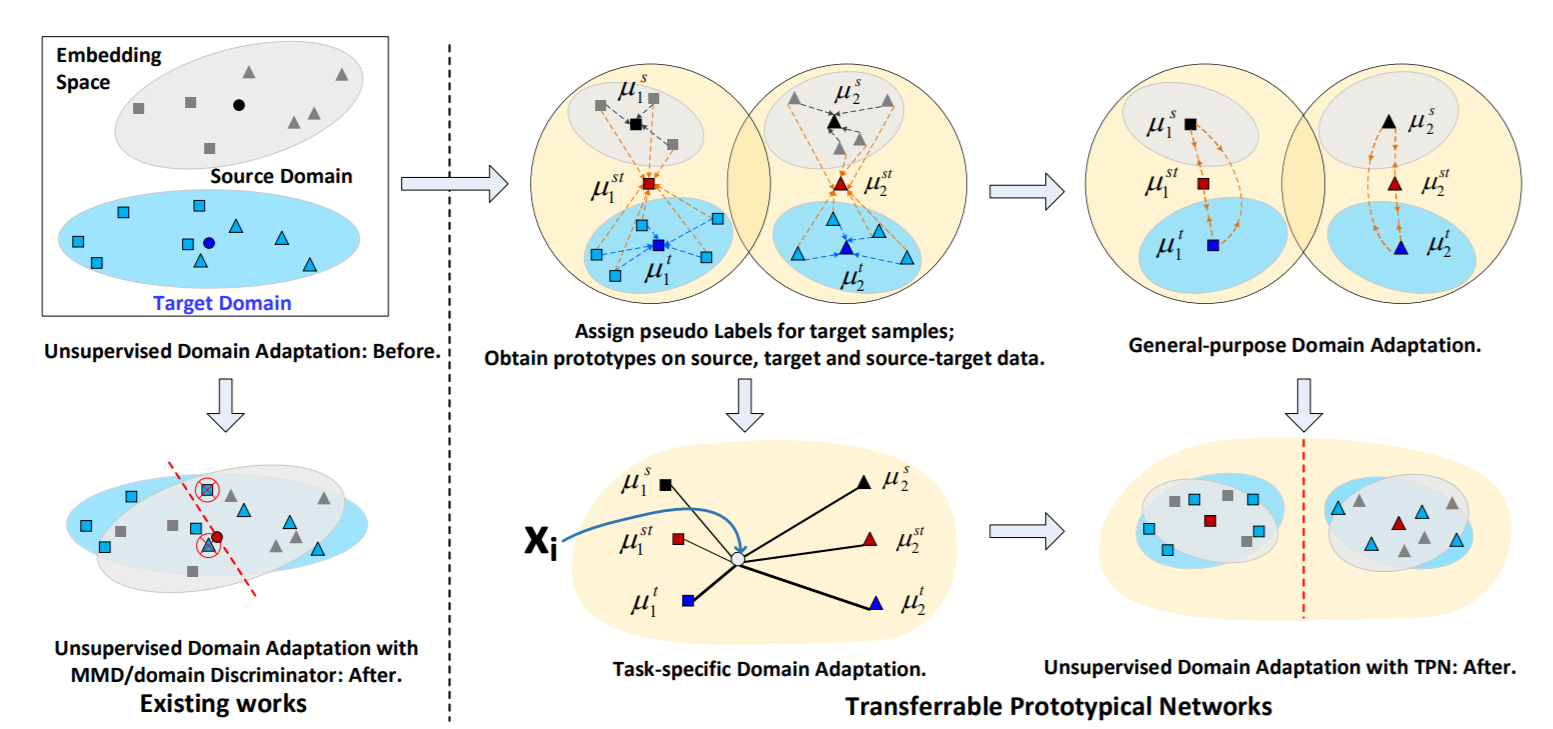}
\caption{The difference between the previous unsupervised domain adaption models like MMD \cite{long2015learning} and domain discriminator \cite{tzeng2015simultaneous}. Compared to these techniques, Transferable Prototypical Network (TPN) can target the scenario of unlabelled data samples by jointly bridging the domain gap and the classifiers are constructed with unlabeled target data and labeled source data.  Image Source: \cite{pan2019transferrable}}
\label{fig:tpn}
\end{figure*}

\textit{Technique using using soft $k$-means:} In this approach, the prototype is looked as a separate cluster center and the refinement process tries to cluster locations to better fit the labelled and unlabelled samples. Once the $(c_k)$ are generated for the labelled samples and clusters are formed based on the distance between the prototypes and the refined prototypes. For the unlabelled samples, they are first partially assigned to the $(c_k)$ clusters based based on the distance between the $(c_k)$ and $(\hat{c_k})$. Finally, refined prototypes are obtained by incorporating the unlabelled samples. This approach is used in cases where the unlabelled image belongs to one of the class in labelled set.

\textit{Technique using using soft $k$-means with distractor class:} The soft $k$-means approach does not perform on the distractor class $(R)$, where the unlabelled samples does not necessarily are from the range of classes. The distractor class images can be harmful when it comes to the $k$-means approach as the $(\hat{c_k})$ prototypes have to be adjusted to the $(c_k)$ clusters. To overcome this, the authors suggest to add an additional cluster which can capture the distractors and avoid the unnecessary population of the $(c_k)$ clusters.

\textit{Technique using using soft $k$-means and masking:} The soft $k$-means with distractor class technique works better for distractor class where all the samples belong to one class. But this is was to simplistic and in real world, it is most unlikely to have a distractor class with samples belonging to just one class. To overcome this, the authors consider that the examples are not within some area of any $(c_k)$ clusters generated from the labelled data samples. This is achieved using the masking procedure. The high-level goal is to mask more the samples which are further away from a prototype and mask less the ones which are closer. Code: \url{https://github.com/xinzheli1217/learning-to-self-train}

\begin{figure*}[t!]
\includegraphics[width=0.9\linewidth]{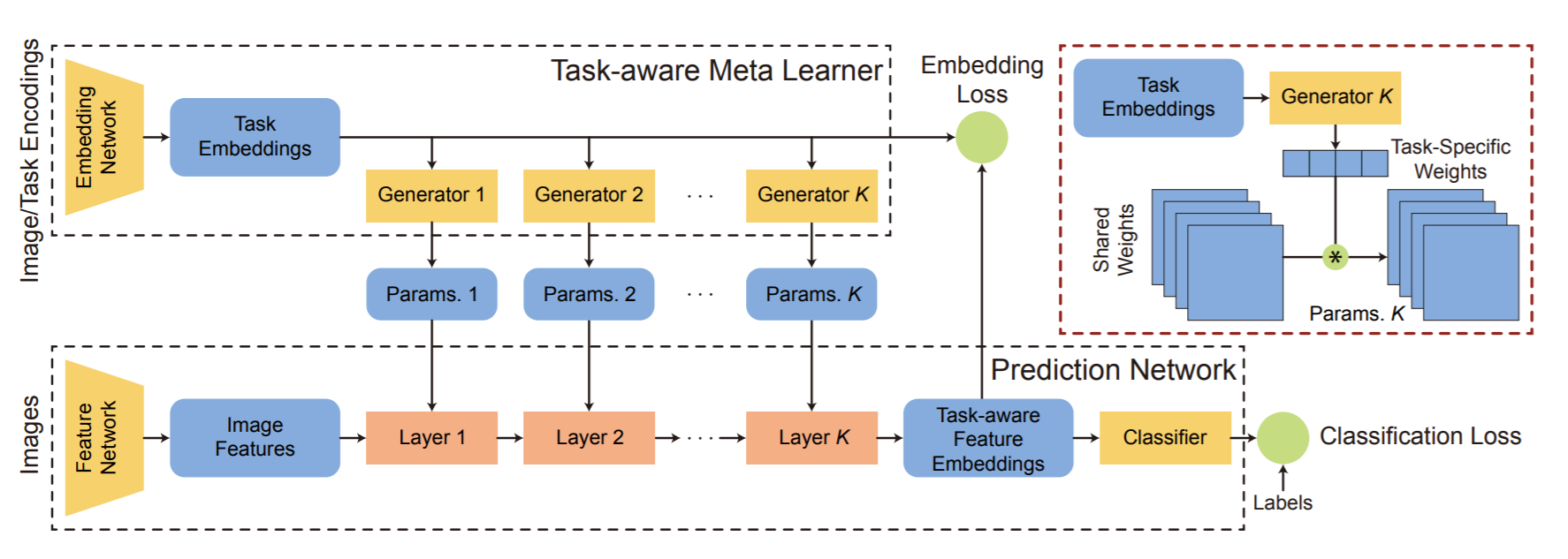}
\caption{The design of the TAFE-Net. TAFE-Net has two modules, meta learner and a prediction network. Depending upon the task, the meta-learner modules learns and produces features for a particular task and the prediction layer adjusts to the new task based on the features generated by meta network. Image Source: \cite{wang2019tafe}}
\label{fig:tafe}
\end{figure*}

\subsubsection{Transferable Prototypical Networks}

The work done by Yingwei et al. \cite{pan2019transferrable} is based on the remoulding the vanilla prototypical network to Transferable Prototypical Network (TPN) which can target the scenario of unlabelled data samples by jointly bridging the domain. The classifiers are constructed with target data which does not have labels along with source data and its respective labels. Initially the transferable prototypical networks classifiers learn from the source data and then directly predict the pseudo labels of the target data which does not have any labels. This results in the generation of two prototypical network based classifiers which are target-only and source-only. TPN training is done to simultaneously reduce the discrepancy at sample level and class level while predicting the correct sample class. They match the prototypes generated from each class and reduce the class-level discrepancy. Also, by enforcing the score distributions over classes of each sample in different domain, the sample-level discrepancy is reduced. 

As illustrated in \autoref{fig:tpn}, firstly, the TPN model allocates a `pseudo' label to each of the target class. This is achieved by matching prototypes of each samples from the target class to the nearest prototype from the source data samples. Afterwards, prototypes are generated based on source-only, target-only and source-target samples. Based on a general purpose adaptation, these prototypes (generated from samples in each domain in all the classes) are pushed to the closest domain in the embedding space. TPN simultaneously aligns the distribution of scores generated by the prototypes of all the samples in various domains. This aligning is performed by doing the task-specific adaptation of the data samples. The entire TPN network is end-to-end trained by reducing the classification loss on the source data along with the general-purpose and task-specific adaptation


\subsubsection{Matching Network}

The work done by Vinyals et al. \cite{vinyals2016matching} is inspired from metric learning techniques \cite{atkeson1997locally, cai2018memory, davis2007information, globerson2006metric}, memory networks \cite{weston2014memory, sukhbaatar2015end}, pointer networks \cite{vinyals2015pointer} and augmented neural networks \cite{santoro2016meta}. The matching network's novelty was two-fold: one, a novel training method tailored for one-shot learning applications; two, introduce a novel network called matching network, which is based on the attention mechanism \cite{cho2014learning, sutskever2014sequence}. Code: \url{https://github.com/AntreasAntoniou/MatchingNetworks}


In matching network, the use of fully differentiable attention mechanism is done to read and/or write from the external memory. The external memory stores important information or knowledge which is pertinent to the task at hand. The essence of matching networks is, without any  modifications to the network it can generate accurate labels for the data samples $D_{test}$. The matching network maps data samples and their respective label in $(S)$, where $(S) = \{(x_n, y_n\}^k$, is mapped to a classifier $c_s(x^\prime)$ which can define a probability distribution over possible output categories $y^\prime$, given an input $x^\prime$ from the $D_{test}$. When $(S)$ is mapped to $c_s(x^\prime)$, the probability is given by:

\begin{equation}
\label{equ:P}
P(y^\prime|x^\prime, S ) = \sum\limits_{n=1}^k a ( x^\prime,x_n)y_n
\end{equation}

where, $a$ is the attention mechanism. So when a small support set $(Q)$ from $D_{test}$ is provided to the one-shot learning model, the distribution of the output classes $y^\prime$ is calculated based on:

\begin{equation}
\label{equ:Pp}
P(y^\prime|x^\prime, Q)
\end{equation}

\subsubsection{Task dependent adaptive metric learning}

The work done by Oreshkin \cite{oreshkin2018tadam} focuses on a novel technique of metric scaling which improves the performance of few-shot applications. Metric training is trying to learn a suitable distance function or a similarly measure (example: cosine or euclidean) as shown above in the work done by Snell et al. \cite{snell2017prototypical} and Vinyals et al. \cite{vinyals2016matching} in prototypical network and matching network, respectively, which is loosely based on the work done by Perez et al. in \cite{perez2017learning, bauer2017discriminative, munkhdalai2017rapid}. They claim that the improvement in the performance of few-shot techniques can be directed key to using different scaling methods and select a particular one based on the softmax from the pool of metrics. They propose a learnable parameter which can make the model understand the best possible metric from the collection of metrics. They also propose to use task conditioning where the embedding generated are not based on a general embedding function but the functions varies with different tasks. Code: \url{https://github.com/ElementAI/TADAM}

\subsubsection{Representative-based metric learning}

The work done by Karlinsky et al. in \cite{karlinsky2019repmet} introduces an effective approach for few-shot object classification and detection  using a technique based on Distance Metric Learning (DML). Their training is inspired on an end-to-end manner where the network simultaneously trains and learns the network parameters, the embedding space, the feature space and the representative vectors. In their work, every class is represented by a mixture model combined with multiple modes. The center of each of these modes is called as a representative vector. Representative vectors will vary along with their respective class. Code: \url{https://github.com/jshtok/RepMet}


\subsubsection{Task-Aware Feature Embedding}

The work done by Wang et al. \cite{wang2019tafe} focuses on the construction of feature embeddings that are set for a particular task. To achieve their goal, they use a novel model called TAFE-Net (Task-Aware Feature Embedding Network) which has two modules or subnetworks: meta learner; followed by a prediction network. Depending upon the task, the meta-learner modules learns and produces features for a particular task and the prediction layer adjusts to the new task based on the features generated by meta network.

As illustrated in \autoref{fig:tafe}, the TAFE-Net generates the TAFEs from the generic image. This is achieved through the meta learner module which is able to generate the feature representation of the different layers in the classification module. The generated weights for these layers are transformed into a task-specific low dimension embedding whereas the high dimension weights are shared among all the global tasks thus reducing the overall complexity. The classifier module used is the same irrespective of the intended task and the input to the classifier are the TAFEs generated by the meta-learner module. Code: \url{https://github.com/ucbdrive/tafe-net}




\subsection{Optimization-Based Techniques}
\label{opti_based}
The techniques discussed in the following section involve the use of an meta-optimizer which can generalize better to novel tasks. An external memory network, Long Short-Term Memory (LSTM) \cite{hochreiter1997long}, a recurrent neural network (RNN), a holistic gradient descent optimizer, etc. are various types of techniques which are used as a meta-optimizer during the initial training phase. The goal of the meta-optimizer is to learns from various tasks and generalize to novel tasks.

\subsubsection{LSTM-based Meta Learner}

The work done by Ravi et al. in \cite{ravi2016optimization} is based on a Long Short-Term Memory (LSTM) network acting as a meta-learner model to generalize and learn the optimal optimization algorithm which can be used to trainer a classifier from another model which has an application towards few-shot regime. Training of most deep neural networks is based on some standard gradient descent algorithm for optimizing the network towards a specific task. 

\begin{equation}
\label{equ:theta_t}
\theta_t = \theta_{t-1} - \alpha_t \nabla_{\theta_{t-1}} \mathcal{L}_t
\end{equation}

where at $t^{th}$ iteration,  $\theta$ is neural network parameters, $\alpha_t$ is the learning rate, $\mathcal{L}_t$ is the loss and $\theta_{t-1} \mathcal{L}_t$ is the gradient of the loss.

The meta-learner LSTM can learn and adapt its rules for training the model. The initial state of the LSTM cell is set to $\theta_t$ and the candidate cell state is set to the gradient of the loss which is $\theta_{t-1} \mathcal{L}_t$ which can determine how valuable the information of the gradient is. They determine the parameters for $n_t$ and $f_t$. The meta-learner takes these values and finds the best possible values which can be used during the training period. $n_t$ and $f_t$ are given by:

\begin{equation}
\label{equ:n_t}
n_t  = \sigma \bigg( W_N \cdot [\theta_{t-1} \mathcal{L}_t, \mathcal{L}_t, \theta_{t-1}, n_{t-1}] + B_N \bigg)
\end{equation}

\begin{equation}
\label{equ:f_t}
f_t =  \sigma \bigg( W_F \cdot [\theta_{t-1} \mathcal{L}_t, \mathcal{L}_t, \theta_{t-1}, f_{t-1}] + B_F \bigg)
\end{equation}

where $W$ indicates the weights and $B$ indicates the biases of the overall model. Based on the $n_t$ and $f_t$ information, the LSTM can learn quickly without diverging. Code: \url{https://github.com/markdtw/meta-learning-lstm-pytorch}

\begin{figure*}[t!]
\centering
\includegraphics[width=0.9\linewidth]{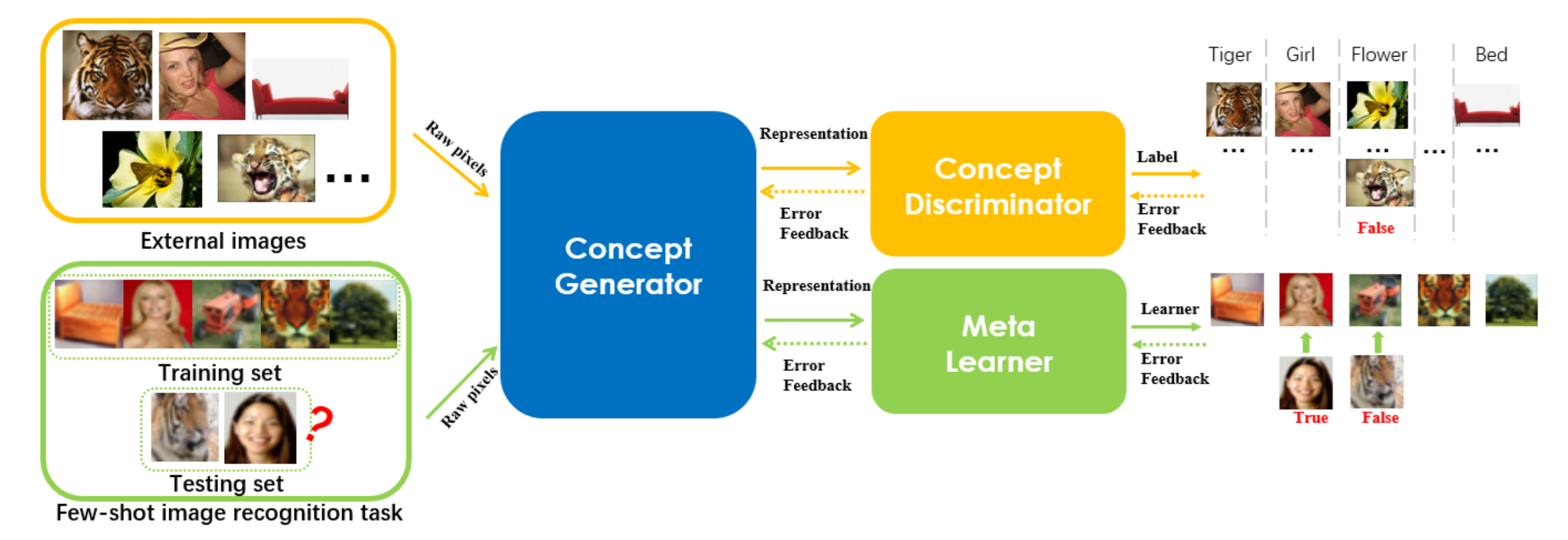}
\caption{The overview of the deep meta-learner. The network consists of three sub-modules: a concept generator $\mathcal{G}$; a meta-learner $\mathcal{T}$; and a concept discriminator $\mathcal{D}$. The main objective is to train the concept generator in parallel with the meta-learner on a series similar tasks, through which the performance of a vanilla meta-learner is enhanced. Image Source: \cite{zhou2018deep}}
\label{fig:dml}
\end{figure*}

\subsubsection{Memory Augmented Networks based Learning}

The work done by Santoro et al. in \cite{santoro2016one} uses a Neural Turing Machine (NTM) \cite{graves2014neural} which they consider as a fully differentiable implementation of Memory Augmented Neural Network (MANN). They use a LSTM as the controller which can communicate with the external memory by using a variety of read and write heads. The speed of information exchange between the model and the external memory is rapid as they external memory stores the vector representation which is been moved in or moved out of the memory because of which NTM is a great option for the application of meta-learning and n-shot predictions. Once the NTM learns how to strategically place the vector representation of the data to be later used for making prediction for the data samples in $D_{test}$. Code: \url{https://github.com/vineetjain96/one-shot-mann}




\subsubsection{Model Agnostic based Meta Learning}

The work done by Finn et al. \cite{finn2017model} proposes to train the weights of a given neural network in a way that, the network can generalize to novel tasks with just few samples. Their approach provided the modern performances on the novel tasks along with quick fine-tuning. Their model has produced good results in the reinforcement learning domain as well, where they achieved quick fine-tuning for the gradient based policies. Their mechanism is able to quickly tune the weights of a model so that it can generalize and adapt quickly to novel tasks. The idea behind their approach was based on the fact that some internal parameters are more transferable than others. For example, in a CNN, the higher levels of convolution can learn features which can be applicable to a variety of tasks irrespective of the intended task. The authors exploit this fact and using a gradient based fine-tuning, they show that the model can rapidly learn and progress onto novel tasks without worrying about overfitting. Algorithm \ref{alg:maml} depicts the overall learning process of the MAML approach. Code: \url{https://github.com/cbfinn/maml}

\begin{algorithm}[b!]
\caption{The MAML (Model-Agnostic Meta Learning) process summarization. Source \cite{finn2017model}}
    \label{alg:maml}
    \begin{algorithmic}
    
    \State \textbf{Requirements:} Task Distribution $p(\mathcal{T})$, setp size $\gamma$, learning rate $\beta$
    \State Initialize random values for $\theta$
    \While {not done}
    \State Perform task batch $\mathcal{T}_i \sim p(\mathcal{T})$
    \For {all $\mathcal{T}_i$}
    \State Evaluate $\nabla_{\theta}\mathcal{L}_{\mathcal{T}_i}(f_{\theta}) $ wrt \textit{k} samples
    \State Calculate the adapted parameters: \\ \hspace{9mm} $\theta^\prime_i = \theta - \gamma \nabla_{\theta}\mathcal{L}_{\mathcal{T}_i}(f_{\theta})$
    \EndFor
    \State Update $\theta \leftarrow \theta - \beta\nabla_{\theta} \sum_{\mathcal{T}_i \sim p(\mathcal{T})} \mathcal{L}_{\mathcal{T}_i}(f^\prime_{\theta}) $
    \EndWhile

    \end{algorithmic}

\end{algorithm}

\begin{figure}[b!]
\centering
\includegraphics[width=0.8\linewidth]{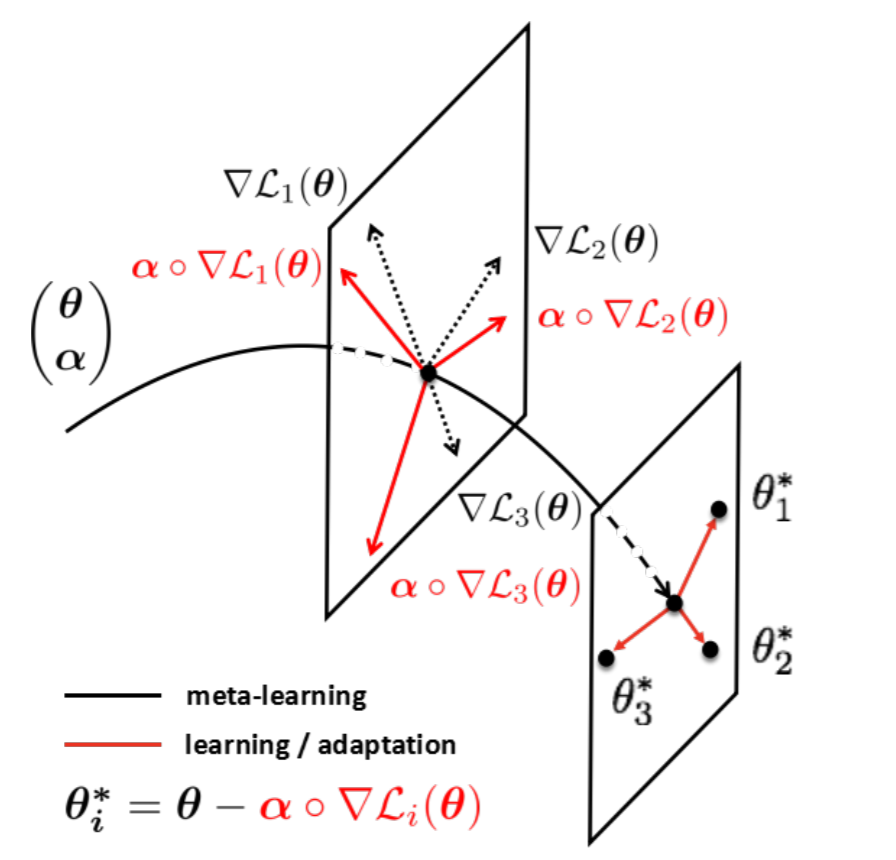}
\caption{The overview of the Meta-SGD learning process. The meta-SGD which can quickly adapt to novel tasks and is applicable for supervised  reinforcement learning domain. Image Source: \cite{li2017meta}}
\label{fig:meta-sgd}
\end{figure}

\subsubsection{Task-Agnostic Meta-Learning}

The work done by Jamal et al. in \cite{jamal2019task} proposes the use of a task-agnostic learning approach. The model, trained initially on a dataset with a range of tasks, can be biased towards few tasks especially when the novel tasks and the trained have some disparities, which can result in the model to be over-performing to few of the tasks that can prevent the meta-learning to learn better to the novel tasks. To overcome this problem, the authors use an unbiased learner able to learn better on the novel tasks. The authors provide two variants for the task-agnostic meta-learner (TAML): Entropy-Maximization/Reduction (EMR-TAML); and Inequality Minimization (IM-TAML). In EMR-TAML, thus avoid the model to overfit to the novel tasks. The authors use a random guess with an equal probability over the predicted labels, which results in biased predictions towards the novel tasks. This is indicated by maximum entropy over the model parameters $(\theta)$ and results in the initial model having a large entropy over the predicted labels. The entropy  is given in \autoref{equ:taml_1}. Alternatively, they also propose to minimize the entropy $(\mathcal{H}_{\mathcal{T}_i} (f_{\theta_i}))$ resulting in higher confidence levels towards the predicted labels after the model parameters are updated from $\theta$ to $\theta_i$ in the process to find the optimum $\theta$. 

\begin{equation}
\label{equ:taml_1}
\mathcal{H}_{\mathcal{T}_i} (f_\theta) = - \mathbb{E}_{x_i \sim P_{\mathcal{T}_i}(x)} \sum_{n = 1}^N \hat{y}_{i,n}log(\hat{y}_{i,n})
\end{equation}

To nullify the biased effect of a model to any particular task, they authors use an approach based on `economic inequality' \cite{marseillecanazei} to measure the amount of task being biased. The approach of economic inequality is inspired from the statistics family, where the loss for each task $\mathcal{T}_i$ of the initial model is looked as an input for the task. Afterwards, the TAML minimizes this loss inequality for multiple tasks, resulting in better meta-learning.

\subsubsection{Meta-SGD}

The work done by Li et al. in \cite{li2017meta} proposes a SGD like optimizer called Meta-SGD which can easily and quickly adapt to novel tasks and is applicable for supervised learning (classification, regression, etc.) and reinforcement learning \cite{duan1611rl2, mishra2017meta, sung2017learning, wang2016learning} domain. Meta-SGD is similar to Meta-Learning LSTM \cite{ravi2016optimization} in terms of easy to implement, conceptually simple, easy to train, and can achieve better performance than Meta-Learning LSTM. Experimental results show that Meta-SGD does indeed outperform MAML approach.

\begin{figure*}[t!]
\centering
\includegraphics[width=0.8\linewidth]{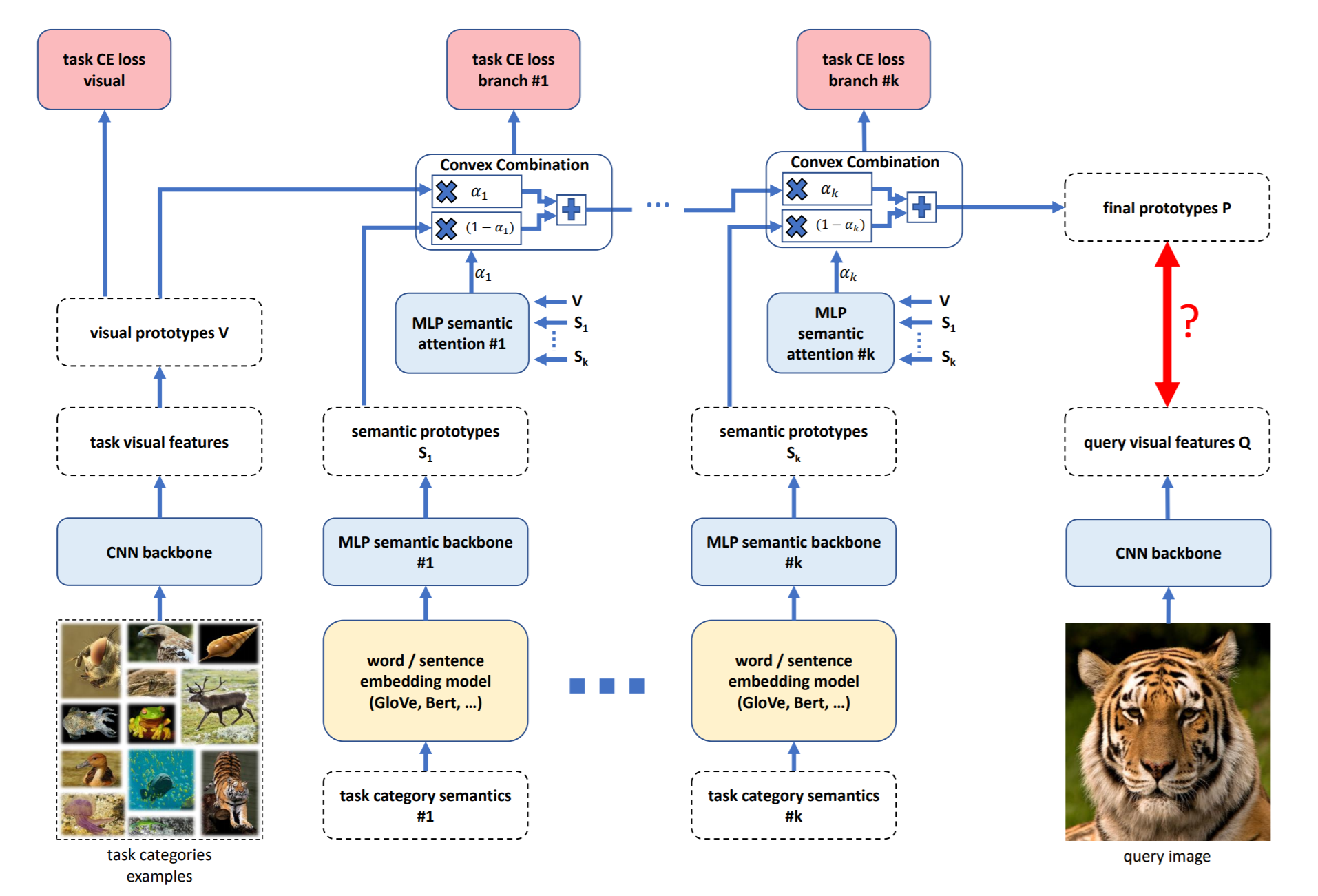}
\caption{The overview of the multiple semantics method where they include various high-level information like natural language, category labels, and attributes. They split the training into two stages: Phase one, where the train the CNN backbone of the network from scratch using the labelled samples.  Image Source: \cite{schwartz2019baby}}
\label{fig:mlp}
\end{figure*}

\autoref{fig:meta-sgd} ilustrates the overall learning procedure of the Meta-SGD. The inspiration is based on a meta-learner, where the meta-learner can generalize from a range of different tasks and can generalize better to a novel task. The learning in Meta-SGD is based on two steps: One, where the meta-learner gradually learns on the different tasks in the meta-space $(\theta, \alpha)$; two, where based on the feedback of the meta-learner the learning approach of the meta-learner is evolved in the learning space. Code: \url{https://github.com/foolyc/Meta-SGD}

\subsubsection{Learning to Learn in the Concept Space}

The work done by Zhou et al. in \cite{zhou2018deep} incorporates the representation power of deep learning into the task of meta learning. As illustrated from \autoref{fig:dml}, their approach includes three sub-modules: a concept generator $\mathcal{G}$; a meta-learner $\mathcal{T}$; and a concept discriminator $\mathcal{D}$. The concept generator is a deep neural network (e.g. ResNet, Inception, VGG Net) which extracts features from the input images, the meta-learner learns on these extracted features and the concept discriminator (e.g SVM, DNN, fully connected layers, etc.) differentiates between the features to recognize different images. The main goal in this approach is to train the concept generator in parallel with the meta-learner on a series similar tasks, which they claim to improve the performance of a vanilla meta-learner. (Deep Meta-Learning) DEML in combination with several other techniques (like Matching Network \cite{vinyals2016matching}, MAML \cite{finn2017model}, Meta-SGD \cite{li2017meta}) has shown to outperform the then best results for the $n$-shot learning problem.

The goal is to minimize joint expectation ($J$), of the meta learning loss $\mathcal{L}_{\mathcal{T}} (\theta_{\mathcal{T}}, \theta_{\mathcal{G}})$ and the discriminator loss $\mathcal{L}_{(x, y)} (\theta_{\mathcal{D}}, \theta_{\mathcal{G}})$ of the labelled samples $(x, y)$ in $D_{train}$

\begin{algorithm}[b!]
\caption{The deep meta learning process summarization. Source \cite{zhou2018deep}}
    \label{alg:dml}
    \begin{algorithmic}
    
    \State \textbf{Input:} Task Distribution $p(\mathcal{T})$, labelled Dataset $D_{train}$, learning rate $\beta$
    \State \textbf{Output:} $\theta_{\mathcal{G}}, \theta_{\mathcal{T}}, \theta_{\mathcal{D}}$
    Initialize $\theta_{\mathcal{G}}, \theta_{\mathcal{T}}, \theta_{\mathcal{D}}$
    \While {not done}
    \State Sample task batch $B_t$
    \State Sample instance batch $B_i$
    \State Compute meta learning loss: $\mathcal{L}_{B_t} (\theta_{\mathcal{T}}, \theta_{\mathcal{G}})$
    \State Discriminator loss: $\mathcal{L}_{B_i} (\theta_{\mathcal{D}}, \theta_{\mathcal{G}})$ 
    \State $(\theta_{\mathcal{G}}, \theta_{\mathcal{D}}, \theta_{\mathcal{T}}) \leftarrow (\theta_{\mathcal{G}}, \theta_{\mathcal{D}}, \theta_{\mathcal{T}})$ 
    \State $-\beta \nabla \bigg[ J(\mathcal{L}_{B_t} (\theta_{\mathcal{T}}, \theta_{\mathcal{G}}), \mathcal{L}_{B_i} (\theta_{\mathcal{D}}, \theta_{\mathcal{G}})) \bigg]$
    \EndWhile

    \end{algorithmic}

\end{algorithm}

\subsubsection{$\Delta$-encoder}

The research done by Schwartz et al. in \cite{schwartz2018delta} proposes the use of an effective and comparatively easy approach for the one-shot and few-shot learning settings where they build upon a modified auto-encoder calling it $\Delta$-encoder. This $\Delta$-encoder can generalize/predict novel tasks based on exposure to few samples from those tasks. The $\Delta$s are the transferable intra-class parameters. The classifier used is able learn how to extract these $\Delta$s, and at the same time also transfers the $\Delta$s towards the prediction of novel samples. Code available at \url{https://github.com/EliSchwartz/DeltaEncoder}





\subsection{Semantic-Based Techniques}
\label{semantic_based}
\begin{table*}[t!]
\normalsize
\centering
\begin{tabular}{l|l|l|l|l|l} \hline
Model                & Technique  & Fine Tune & \multicolumn{3}{c}{5-Way Accuracy (\%)}                                \\ \hline
                     &              &           & 1-shot                        & 5-shot                        & 10-shot \\ \hline
MEPS \cite{chu2019spot} & Data Augmentation  & N & 97.56 $\pm$ 0.31 & 99.65 $\pm$ 0.06 & - \\ \hline
Relation Network \cite{sung2018learning}     & Embedding    & N         & 99.6 $\pm$  0.2                       & 99.8 $\pm$  0.1                           & -       \\
Matching Network \cite{vinyals2016matching}     & Embedding    & Y         & 97.9                          & 98.7                          & -       \\
Prototypical Network \cite{snell2017prototypical} & Embedding    & N         & 98.8                          & 99.7                          & -       \\ \hline
MANN \cite{santoro2016meta}                & Optimization & N         & 96.4                          & 94.9                          & 98.1    \\
MAML \cite{finn2017model}                 & Optimization  & N         & 98.7 $\pm$ 0.4   & 99.9 $\pm$ 0.1   & -       \\
TAML (entropy) \cite{jamal2019task}      & Optimization & N         & 99.23 $\pm$ 0.35 & 99.71 $\pm$ 0.1  & -       \\
TAML (EMR-TAML)  \cite{jamal2019task}          & Optimization & N         & 99.1 $\pm$ 0.36  & 99.6 $\pm$ 0.1   & -       \\
TAML (IM-TAML)  \cite{jamal2019task}       & Optimization & N         & 99.47 $\pm$ 0.25 & 99.83 $\pm$ 0.09 & -       \\
Meta- SGD \cite{li2017meta}           & Optimization & N         & 99.53 $\pm$ 0.26 & 99.93 $\pm$ 0.09 & -       \\ \hline
\end{tabular}
\caption{Comparison Results of various techniques on Omniglot Dataset \cite{lake2015human, lake2019omniglot}. $\pm$ indicated 95\% confidence intervals over different tasks.} 
\label{table:results_omniglot}
\end{table*}

\begin{table*}[t!]
\normalsize
\centering
\begin{tabular}{l|l|l|l|l|l} \hline
Model                           & Technique    & Fine Tune & \multicolumn{3}{c}{5- Way Accuracy (\%)}                                \\ \hline
                                &              &           & 1-shot                        & 5-shot                        & 10-shot \\ \hline
SGM  \cite{hariharan2017low} & Data Augmentation & N & 45.1 & 72.7 & 79.1 \\ 
SalNet \cite{zhang2019few}      &  Data Augmentation & N & 57.45 $\pm$0.88 & 72.01 $\pm$ 0.67 & -  \\
MEPS \cite{chu2019spot} & Data Augmentation & N & 51.03 $\pm$ 0.78 & 67.96 $\pm$ 0.71 & - \\
PMN \cite{wang2018low} & Data Augmentation & N & 57.6 & 71.9 & 75.2 \\ 
DeMe-Net  \cite{chen2019image} & Data Augmentation & N & 59.14 $\pm$ 0.86 & 74.63 $\pm$ 0.74 & - \\ \hline
Relation Network \cite{sung2018learning}               & Embedding     & N         & 50.44 $\pm$ 0.82 & 65.32 $\pm$ 0.70 & -       \\
Matching Network \cite{vinyals2016matching}               & Embedding     & Y         & 43.56 $\pm$ 0.84 & 55.31 $\pm$ 0.73 & -       \\
Prototypical Network \cite{snell2017prototypical}           & Embedding     & N         & 49.42 $\pm$ 0.78 & 68.20 $\pm$ 0.66 & -       \\
Prototypical Network + BF \cite{wertheimer2019few}      & Embedding     & N         & 47.67 $\pm$ 0.31 & 65.2 $\pm$ 0.29  & -       \\
Prototypical Network + fsL \cite{wertheimer2019few}     & Embedding    & N         & 51.1 $\pm$ 0.3   & 67.85 $\pm$ 0.29 & -       \\
Prototypical Network + fsL + CP \cite{wertheimer2019few} & Embedding    & N         & 49.64 $\pm$ 0.31 & 69.45 $\pm$ 0.28 & -       \\
Soft k-means \cite{ren2018meta}                    & Embedding    & N         & 50.09 $\pm$ 0.45 & 64.59 $\pm$ 0.28 & -       \\
Soft k-means + Cluster \cite{ren2018meta}         & Embedding   & N         & 49.03 $\pm$ 0.24 & 63.08 $\pm$ 0.18 & -       \\
Masked k-means \cite{ren2018meta}                 & Embedding    & N         & 50.41 $\pm$ 0.31 & 64.39 $\pm$ 0.24 & -       \\
TADAM  \cite{oreshkin2018tadam}                         & Embedding    & Y         & 58.5                          & 76.7                          & 80.8    \\ \hline
Meta-Learning LSTM  \cite{ravi2016optimization}            & Optimization  & N         & 43.44 $\pm$ 0.77 & 60.60 $\pm$ 0.71 & -       \\
MAML \cite{finn2017model}                           & Optimization  & N         & 48.7 $\pm$ 1.84  & 63.00 $\pm$ 0.92 & -       \\
Meta-SGD \cite{li2017meta}                        & Optimization  & N         & 50.47 $\pm$ 1.87 & 64.03 $\pm$ 0.94 & -       \\
DEML + Matching Network \cite{zhou2018deep}         & Optimization  & N         & 55.84 $\pm$ 0.94 & 59.88 $\pm$ 0.73 & -       \\
DEML + MAML \cite{zhou2018deep}                     & Optimization  & N         & 53.71 $\pm$ 0.89 & 68.13 $\pm$ 0.77 & -       \\
DEML + Meta-SGD \cite{zhou2018deep}                 & Optimization  & N         & 58.49 $\pm$ 0.91 & 71.28 $\pm$ 0.69 & -       \\
$\Delta$-encoder  \cite{schwartz2018delta}                           & Optimization & Y         & 59.9                       & 69.7                         & - \\\hline
Multiple Semantics \cite{schwartz2019baby} & Semantics & N & \textbf{67.3} & \textbf{82.1} & - \\ \hline

\textbf{4-5 years old child} & Human Performance & & 70 & - & -  \\
\textbf{Adult} & Human Performance & & 99 & - & - \\ \hline

\end{tabular}
\caption{Comparison Results of various techniques on \textit{Mini}ImageNet Dataset \cite{vinyals2016matching}. $\pm$ indicated 95\% confidence intervals over different tasks.}
\label{table:results_miniimagenet}
\end{table*}

This section, discusses the semantic-based techniques, in which the semantics is included along with the samples to learn and generalize better to novel tasks. Semantic-based techniques are used more popularly for zero-shot learning (ZSL) settings. The inspiration behind this approach is that, often an adult is accompanied, when pointing out a new thing to a child, to make the association. The information or pointing out of an adult can be compared as a semantic knowledge which helps the child learn novel classes. In traditional zero-shot learning settings, either, the semantics and the data are mapped to a common embedding space \cite{zhang2015zero}, or, the semantics are mapped to the data or vice-versa \cite{zhang2017learning, frome2013devise}. 

\subsubsection{Learning with Multiple Semantics}

Schwartz \cite{schwartz2019baby} presents an incremental work to few-shot learning techniques discussed in \cite{xing2019adaptive}, where they incorporate additional semantic information that can help the learning process to be more effective. They propose that by combining multiple high-level semantics like natural language description, category labels, and attributes. They split the training into two stages: Phase one, where the train the CNN backbone of the network from scratch using the labelled samples. They, however, believe that a pre-trained task specific CNN model fine-tuned to the labelled dataset gives much better performance rather than training it from scratch; Phase two, where the linear classifier is replaced by MLP (multi-layer Perceptron), freezing all the previous layers. The MLP is able to generate `semantic prototypes' which are then added up to the semantic branches as shown in \autoref{fig:mlp}.

\subsubsection{Learning via Aligned Variational Autoencoders (VAE)}

The work done by Schonfeld et al. \cite{schonfeld2019generalized} is incremental to the feature generation technique where a model shares a latent space of the image embeddings and the respective class embeddings. The shared latent space is learned by variational autoencoders (VAE) \cite{doersch2016tutorial} which are modality specific. They evaluated the model on several benchmark datasets and claimed that their model is state-of-the-art for generalized zero-shot learning and few-shot learning techniques. Code: \url{https://github.com/edgarschnfld/CADA-VAE-PyTorch}\cite{schonfeld2019generalized}

\subsubsection{Learning by Knowledge Transfer With Class Hierarchy}

Li et al. in \cite{li2019large} propose a large-scale model where the learnt features are transferable between the class hierarchy and can encode the semantic relation between the source and target class, respectively. They claim that their model is able to outperform the then state-of-the-art approach on a large scale zero-shot learning problem and can also be extended to few-shot learning applications. In high level, the prior knowledge they use is the semantic relation between the source classes and target classes and they claim that by transferring the feature embedding of the source class can help in predicting the target class. Code: \url{https://github.com/tiangeluo/fsl-hierarchy}

\section{Discussion and Future Direction}
\label{discussion}

This section presents the different techniques used and compare their performance on various benchmarks and datasets. For testing of these techniques, the two most commonly used benchmark datasets: Omniglot  and \textit{Mini}ImageNet. Omniglot dataset contain 1623 various handwritten characters from various alphabets. Amazon's Mechanical Turk was used by 20 different people to online draw these 1623 various characters. Omniglot is similar to MNIST dataset with respect to the complexity of the images.  On the other hand, the much complicated, \textit{Mini}ImageNet is a subset of the ImageNet dataset \cite{russakovsky2015imagenet} \footnote{Note: Complicated in the terms how tough it is for the model to generalize to the dataset and provide state-of-the-art performance}. The \textit{Mini}ImageNet dataset contains images of size $84 \times 84$ which belong to randomly selected 100 output categories, with each category having 600 images, i.e. a total of 6000 images. The standard training procedure in a 5-way settings is to have 1 or 5 samples (or it can be any number of samples in between 1 and 10) for each of the 5 output classes in the support set. ( therefore it is called as 5-way) Out of the 100, 64 categories are used for training, 16 for validation whereas 20 are used for testing. 

\autoref{table:results_omniglot} depict the performance of various techniques discussed on the omniglot dataset. The dataset been relative simple, majority of the models are able to achieve high accuracy for 1-shot and 5-shot settings. Even though Data Augmentation and Embedding (or metric) learning achieved high accuracies, the optimization based techniques are a clear winner with an accuracy of 99.53\% using the Meta-SGD algorithm \cite{li2017meta}. Few of the optimization-based techniques have a broader application scope and can be used in reinforcement settings as well including the Meta-SGD algorithm. 

\autoref{table:results_miniimagenet} highlights the performance of various techniques discussed on the \textit{Mini}ImageNet dataset \footnote{Experiments for few of the techniques described in \autoref{semantic_based} were performed on different datasets and therefore have been omitted from the above comparison table}. As evidently seen, there is a considerable drop in the performance of the techniques discussed compared to their performance on the omniglot dataset. Techniques like Matching Network, MAML, Meta-SGD have 95\% accuracy on the omniglot dataset whereas these same models have significantly lower accuracies on the \textit{Mini}ImageNet dataset. A major reason for this drop is that the images in the \textit{Mini}ImageNet dataset are much more complicated in terms of the contextual meaning, rich source of information, etc as compared to the omniglot dataset where the images are just characters. The data augmentation, embedding, optimization approach have 1-shot accuracies around 55\%, even though there is a significant rise in the accuracies for the 5-shot settings, accuracies in the 70\% ballpark. The trend is, more the number of samples, the better the accuracy. In order to make our model learn from as less samples as possible, we focus on the 1-shot accuracies. The state-of-the-art accuracy is achieved through using semantic as an additional source of information along with the data to make the model understand novel categories.


The present state-of-the-art is still deficient when compared to a 4-5 years old preschooler's performance, indicating a substantial scope of improvement in these techniques in the near future to compete or match with an adult human being's performance. A hybrid model which can exploit the advantages of various techniques to boost performance can be the next course of action in the few shot meta learning domain. A hybrid model; a cross-modal implementation which incorporates the semantic information along with the data-augmentation and/or embedding techniques with reference to results in \autoref{table:results_miniimagenet}. An observation at the results generated, acknowledges that using semantic information achieves superiror results in-comparison to data-augmentation or embedding techniques. However,the sole use of semantic based approach still lacks performance quality when compared to that of a human. Incorporating advanced techniques or models like attention mechanism \cite{mnih2014recurrent, bahdanau2014neural}, self-attention mechanism, transformers \cite{vaswani2017attention} or a variation of variational auto encoders (VAEs) like beta-VAE \cite{higgins2016beta}, VQ-VAE \cite{van2017neural}, VQ-VAE-2 \cite{razavi2019generating} and TD-VAE \cite{gregor2018temporal} that can better generate low dimension semantic information for better generalization of limited samples and thus can perform better on novel categories along with an improved distance function (euclidean or cosine) which can robustly classify or cluster the low dimensional embeddings.

\section{Acknowledgement}
\label{Ack}

The authors gratefully acknowledge use of the services of Jetstream cloud, funded by NSF award 1445604 and partly supported by Open Cloud Institute (OCI) at University of Texas at San Antonio.

\section{Conclusion}
\label{conclusion}
With the availability of enough data samples and their respective labels, deep learning models can yield better performance by generalizing well to such tasks, but fail where the model has to learn from limited samples. Representation based learning method like few shot learning, meta learning are used improve the ability of the model to learn from limited samples. In this survey, we put forth an investigation on the finding and existing techniques on  few shot meta learning techniques for supervised learning in the computer vision domain. We highlighted the fact that why it is imperative to research on techniques where the model needs to generalize well based on limited data samples and the prior knowledge.  We classified the techniques into four main categories based on their approach towards solving the few-shot learning problem. We analyzed the performance of these techniques on two benchmarks, Omniglot and \textit{Mini}Imagenet dataset and provided a brief discussion regarding their performance and the ideal settings for those techniques along with a potential future direction for research towards matching or outperforming humans.

\bibliographystyle{IEEEtran}
\bibliography{main}

%

\begin{IEEEbiography}[{\includegraphics[width=1in,height=1.25in,clip,keepaspectratio]{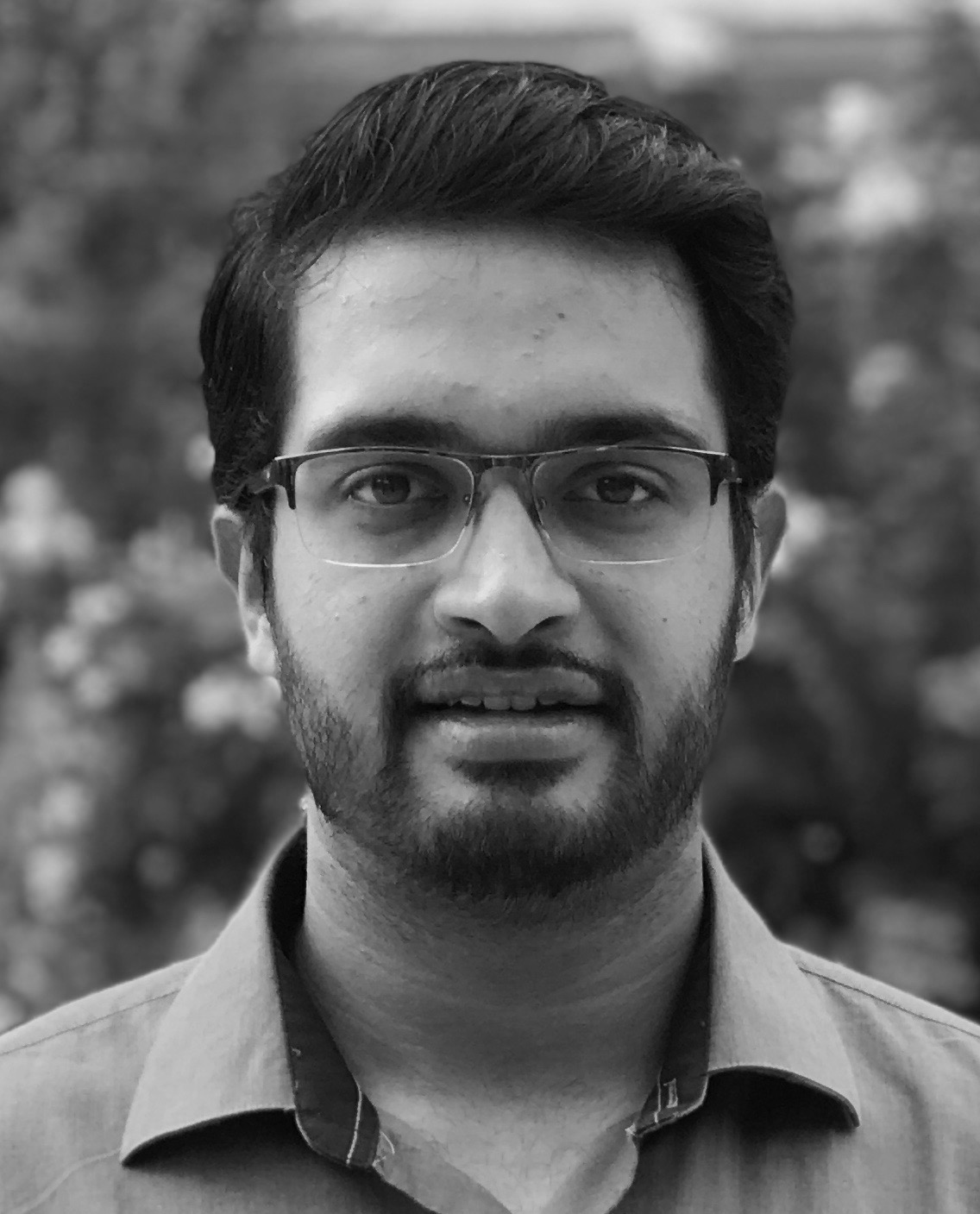}}]{Nihar Bendre}
currently pursuing his Doctor of Philosophy (PhD) Degree in Electrical Engineering from University of Texas at San Antonio, He is dedicatedly pursuing research in the computer vision field of Machine/Deep Learning working in the Secure AI and Autonomy Lab. He finished his Master in Electrical Engineering from University of Texas at San Antonio. Nihar came to USA in 2013 after finishing his Bachelor’s in Engineering degree from University of Pune (Pune, India). His interests are in keeping abreast with the recent technological advancement and in his part time passionately plays cricket with a local club in San Antonio.
\end{IEEEbiography}

\begin{IEEEbiography}
[{\includegraphics[width=1in,height=1.3in,clip,keepaspectratio]{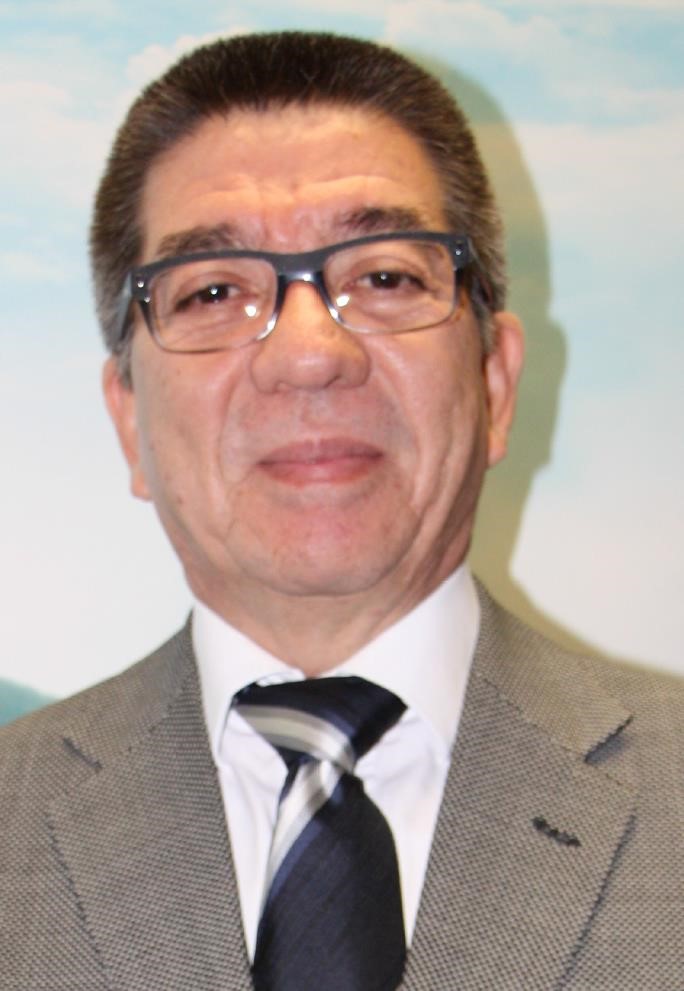}}]{Hugo Terashima-Mar\'in}
holds a BSc in Computational Systems from Tecnológico de Monterrey, Campus Monterrey in 1982; MSc in Computer Science from University of Oklahoma in 1987; MSc in Information Technology and Knowledge-based Systems from University of Edinburgh in 1994; and PhD in Informatics from Tecnológico de Monterrey, Campus Monterrey in 1998.

Dr. Terashima-Mar\'in is a Full Professor at the School of Engineering and Sciences, the Leader of the Research Group with Strategic focus in Intelligent Systems and Director of the Graduate Program in Computer Science. He is a member of the National System of Researchers, the Mexican Academy of Sciences, and the Mexican Academy of Computing.  He is a Senior Member of the IEEE. His research areas are computational intelligence, heuristics, metaheuristics and hyper-heuristics for combinatorial optimization, characterization of problems and algorithms, constraint handling and applications of artificial Intelligence and machine learning. He has been principal investigator of various projects for industry and CONACyT.  Dr. Terashima-Mar\'in has current collaboration with research groups in the University of Nottingham, University of Stirling, University of Edinburgh-Napier, University of Texas-San Antonio, the University Andrés Bello in Santiago de Chile, and the Chinese Academy of Sciences.  He has published more than 90 research articles in international journals and conferences. He has supervised 5 PhD dissertations and 29 Master Thesis. In the past, he has been Director of the MSc in Intelligent Systems, PhD in Artificial Intelligence, PhD in Information Technology and Communications, the PhD Programs, and Graduate Programs at Tecnológico de Monterrey, Campus Monterrey.
\end{IEEEbiography}

\begin{IEEEbiography}
[{\includegraphics[width=1in,height=1.3in,clip,keepaspectratio]{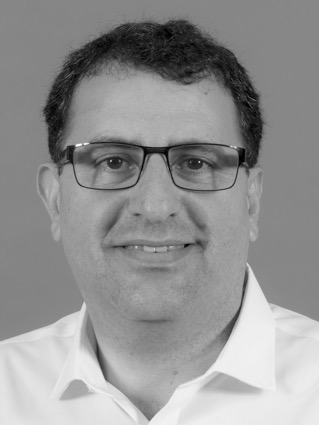}}]{Peyman Najafirad}
is a co-founder and Associate Director of the Open Cloud Institute (OCI), and an Associate Professor with the Information Systems and Cyber Security Department at the University of Texas at San Antonio. He received his first B.S. degree from Sharif University of Technology in Computer Engineering in 1994, his 1st master in artificial intelligence from the Tehran Polytechnic, his 2nd master in computer science from the University of Texas at San Antonio (Magna Cum Laude) in 1999, and his Ph.D. in electrical and computer engineering from the University of Texas at San Antonio.
He was a recipient of the Most Outstanding Graduate Student in the College of Engineering, 2016, earned the Rackspace Innovation Mentor Program Award for establishing Rackspace patent community board structure and mentoring employees (2012), earned the Dell Corporation Company Excellence (ACE) Award  for exceptional performance and innovative product research and development contributions (2007), and earned the Dell Inventor Milestone Award, Top 3 Dell Inventor of the year (2005). He holds 15 U.S. patents on cyber infrastructure, cloud computing, and big data analytics with over 300 product citations by top fortune 500 leading technology companies such as Amazon, Microsoft, IBM, Cisco, Amazon Technologies, HP, and VMware. He has advised over 200 companies on cloud computing and data analytics with over 50 keynote presentations. High performance cloud group chair at the Cloud Advisory Council (CAC), OpenStack Foundation Member (the \#1 open source cloud software), San Antonio Tech Bloc Founding Member, and Children’s Hospital of San Antonio Foundation board member.
\end{IEEEbiography}









\end{document}